\documentclass[letterpaper]{article}
\usepackage[preprint]{aaai2027}
\usepackage[hyphens]{url}
\usepackage{graphicx}
\usepackage{natbib}
\usepackage{caption}
\usepackage{algorithm}
\usepackage{algorithmic}
\usepackage{booktabs}
\usepackage{amsmath}
\usepackage{amssymb}
\usepackage[table]{xcolor}
\usepackage[scr=boondox]{mathalpha}

\title{TimeSteer: Inference-Time Speech Scheduling in Joint Audio-Visual Diffusion Models}
\author{
    Chao Zhou, Yiling Chen, Qi Chu, Tao Gong, Nenghai Yu, Tianyi Wei
}
\affiliations{}

\begin{document}

\maketitle

\begin{abstract}
Although pretrained joint audio-visual diffusion models offer rich control over \emph{what} to generate, they provide no explicit control over \emph{when} an utterance should occur. To address this, we study \emph{inference-time speech scheduling}, a novel task that places coupled speech and visual articulation within user-specified begin--end intervals without finetuning the backbone model.
We uncover two intrinsic properties of the denoising process that enable this task. First, a timing-sensitive text-to-audio cross-attention head exposes each utterance's model-implied source span along the latent timeline. Second, the predicted clean latent already organizes coupled speech and visual articulation, allowing their temporal placement to be edited without regenerating the content. Building on these discoveries, we propose \textbf{TimeSteer}, a training-free framework that localizes each utterance's source span through \textbf{Source Span Localization} and transfers the associated audio-visual latent content from the source interval to the specified target interval through \textbf{Region-Aware Latent Remapping}. We further introduce \textbf{SpeechShift}, the first benchmark for interval-level speech scheduling in joint audio-visual generation. Experiments across two representative backbones show that TimeSteer substantially improves interval controllability over training-free baselines while maintaining competitive overall generation quality.
\end{abstract}

\section{Introduction}

\begin{figure*}[t]
  \centering
  \includegraphics[width=\textwidth]{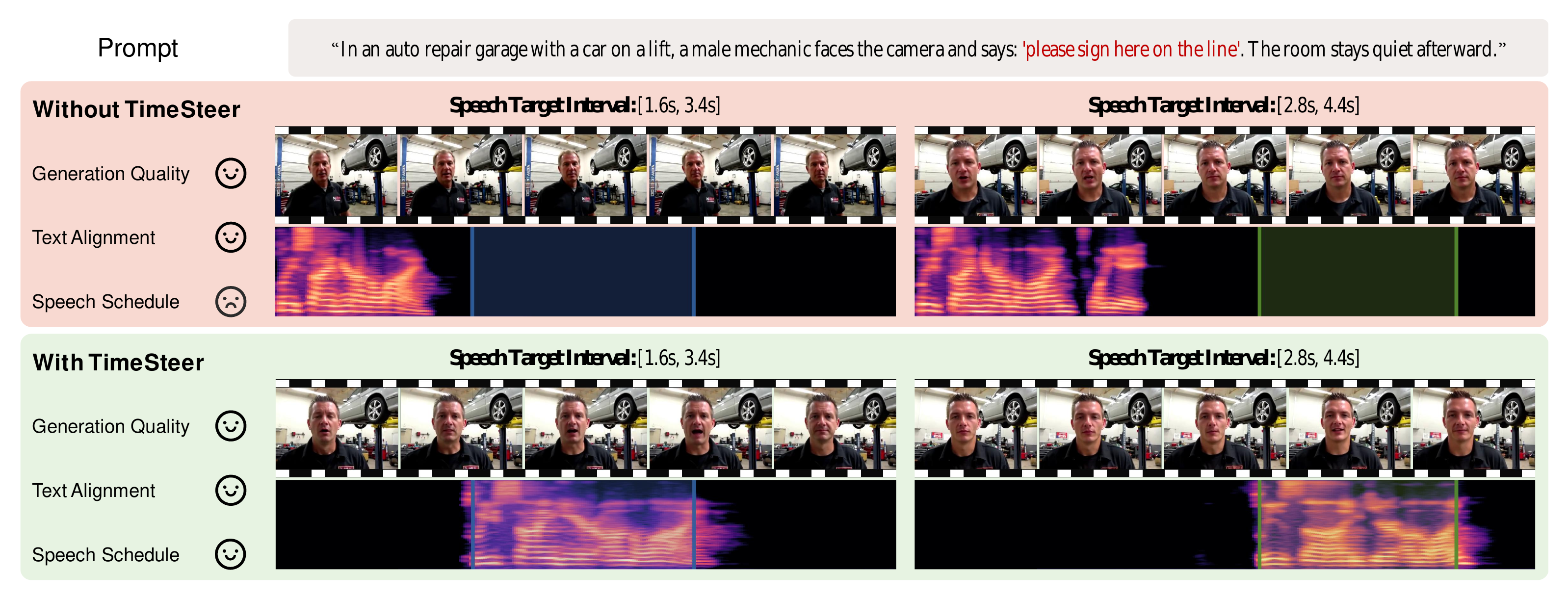}
  \caption{\textbf{Qualitative speech scheduling.} Without TimeSteer, a timing instruction of the form \emph{“from [start] to [end]”} is appended to the original prompt, yet the speech remains near the video onset. TimeSteer places the utterance within any user-specified target interval while preserving generation quality and text alignment. Shading marks the target intervals; frames are sampled at 1\,fps with aligned mel-spectrograms below.}
  \label{fig:teaser}
\end{figure*}

Recent joint audio-visual diffusion models synthesize temporally coupled speech and video in a single generation pass, enabling practical text-to-audio-visual (T2AV) generation.
Systems such as LTX-2~\cite{hacohen2025ltx2} and daVinci-MagiHuman~\cite{chern2026speed} accept diverse conditioning signals, including open-ended text, optional visual inputs, and speech-centric controls such as reference timbre, while producing natural prosody and synchronized lip motion.
These advances provide increasingly rich control over \emph{what} is generated, yet offer almost no control over \emph{when} an utterance begins and ends.

This missing temporal controllability limits scriptable audio-visual generation.
Applications such as filmmaking, interactive agents, and game engines often require dialogue to occupy prescribed begin--end intervals, whether to follow visual events, coordinate multiple speakers, or insert deliberate pauses.
Such interval constraints specify not only utterance onset and offset, but also the effective speaking rate required for each sentence to naturally fit its assigned duration.
However, current joint generators provide no explicit mechanism for enforcing these temporal constraints: speech timing emerges implicitly during denoising, making utterance placement difficult to predict or reproduce across random seeds.
We call this capability \emph{inference-time speech scheduling}: given one or more quoted utterances, generate coupled audio-visual speech that places each utterance within any user-specified begin--end interval, without any retraining or finetuning.

Despite its practical importance, inference-time speech scheduling has received little attention.
Existing T2AV benchmarks such as JavisBench~\cite{wang2025javisdit} and AVGen-Bench~\cite{liu2026avgenbench} evaluate synchronization and semantic fidelity, but not interval-level speech controllability.
Related generation methods either control the timing of isolated acoustic events in text-to-audio generation~\cite{xu2024picoaudio,wang2025controlaudio,zhang2025freeaudio} or synchronize video to externally provided speech~\cite{weng2025mtv,lee2025syncphony}.
The central question therefore remains open: can a pretrained joint generator directly generate coupled speech and visual articulation within user-specified begin--end intervals entirely at inference time?

We answer this question affirmatively by uncovering two intrinsic properties of the pretrained generator's denoising process. First, speech timing emerges internally well before decoding and remains traceable as generation proceeds. In particular, a timing-sensitive cross-attention head naturally exposes the source span of each utterance along the latent timeline. Second, the model's learned audio-visual prior already couples each utterance with its corresponding visual articulation in the predicted clean latent, so localizing its source span is sufficient to jointly reschedule both modalities without regenerating their content.

Motivated by these insights, we propose \textbf{TimeSteer}, a training-free framework that operates during denoising in two successive stages: \emph{Source Span Localization} and \emph{Region-Aware Latent Remapping}. First, Source Span Localization aggregates the responses of a timing-sensitive text-to-audio cross-attention head over the text tokens corresponding to each utterance, thereby estimating its source span without decoding any intermediate predictions. Then, Region-Aware Latent Remapping reorganizes the predicted clean latent to transfer the corresponding joint audio-visual latent content from the localized source span to the target interval. Because a source span and its target interval may differ in duration, edited speech regions and the surrounding non-speech gaps require distinct treatment: the former must adapt the speaking rate to match the user-specified duration, whereas the latter should preserve the original temporal progression as much as possible while smoothly absorbing transitions between speech intervals. Accordingly, we introduce \textbf{Minimal-Distortion Remapping}, which applies an affine map within each speech interval, and \textbf{Curvature Bridging}, which smoothly interpolates the read map across non-speech gaps. Figure~\ref{fig:teaser} shows that TimeSteer achieves precise speech relocation to any user-specified target interval while preserving synchronized audio-visual generation.

Evaluating inference-time speech scheduling requires dedicated benchmarks.
Existing T2AV evaluations assess synchronization and perceptual quality but provide neither target intervals nor metrics for interval-level controllability.
We therefore introduce \textbf{SpeechShift}, the first benchmark designed for this task.
SpeechShift contains 400 prompts across 102 scenes and 600 utterance-level scheduling targets, spanning four combinations of speaker and utterance counts together with diverse acoustic interference and action-coupling conditions. To quantify controllability, SpeechShift measures how closely each realized speech interval matches its target in both boundary timing and temporal overlap. It also evaluates whether this control preserves the quality and synchronization of the generated audio-visual content.

Taken together, TimeSteer and SpeechShift provide a method and benchmark for inference-time speech scheduling. Experiments across two joint audio-visual diffusion backbones show that TimeSteer substantially improves interval controllability over training-free baselines while maintaining competitive generation quality.

Our contributions are:
\begin{itemize}
\item \textbf{Task.} We introduce inference-time speech scheduling, a new task of placing each generated utterance and its coupled visual articulation within any user-specified target interval, without any retraining or finetuning.
\item \textbf{Discovery.} We uncover two intrinsic properties of joint audio-visual denoising: cross-attention exposes the model-implied source span of each utterance, while the predicted clean latent makes the temporal placement of coupled speech and visual articulation directly editable.
\item \textbf{Method.} We propose \textbf{TimeSteer}, a training-free framework that first localizes each utterance's source span and then transfers its joint audio-visual latent content to the target interval without regenerating it. Its region-aware remapping combines uniform temporal scaling within utterances with smooth transitions across non-speech gaps.
\item \textbf{Benchmark.} We introduce \textbf{SpeechShift}, the first benchmark for interval-level speech scheduling in joint audio-visual generation, comprising 400 prompts, 600 utterance-level targets, and 102 scenes. Its evaluation jointly measures interval controllability and the preservation of overall generation quality.
\end{itemize}

\section{Related Work}
\label{sec:related-work}

\subsection{Joint Audio-Visual Generation}

Joint audio-visual generation has evolved from modality-specific fusion toward large-scale diffusion transformers that synthesize sound and video in a unified process. Early methods such as MM-Diffusion and SyncFlow introduced dedicated routing and synchronization mechanisms for cross-modal interaction~\cite{ruan2023mmdiffusion,wang2024syncflow}. Recent work has scaled this paradigm to larger, more integrated architectures~\cite{polyak2024moviegen,xing2024seeing,wang2025animate,li2025harmony,low2025ovi,nava2026,wang2026mova,chen2025unison}. For example, LTX-2 couples separate audio and video streams through cross-attention, whereas daVinci-MagiHuman processes all modality tokens within a unified attention architecture~\cite{hacohen2025ltx2,chern2026speed}.

Building on this progress, recent extensions introduce reference conditioning, alignment objectives, and preference optimization to improve content control and cross-modal consistency~\cite{zhang2026mmcontrol,liu2026syncdpo,ma2026itsjav}. These methods broaden control over what is generated, but do not expose when individual utterances should occur. Existing evaluations follow the same emphasis: JavisBench and AVGen-Bench measure semantic fidelity, perceptual quality, and synchronization, yet provide neither target speech intervals nor metrics for inference-time speech scheduling~\cite{wang2025javisdit,liu2026avgenbench}. As a result, placing model-generated speech and its corresponding visual articulation within user-specified intervals has received little direct attention in joint audio-visual generation.

\subsection{Temporal Control in Audio and Video}

Temporal control has been extensively studied in modality-specific audio and video generation, enabling acoustic events and visual motions to follow explicit timing conditions. In text-to-audio generation, methods build on large audio models~\cite{kreuk2023audiogen,liu2023audioldm,huang2024makeanaudio2} and introduce event timelines, timestamp conditioning, compositional controls, or inference-time attention manipulation to place acoustic events at desired times~\cite{huang2024audiocomposer,xu2024picoaudio,li2025picoaudio2,wang2025controlaudio,zhang2025freeaudio}. Text-to-video methods similarly control temporal evolution through motion trajectories and frame-level conditions~\cite{huang2025tempocontrol,ruan2023aadiff,yariv2023tempotokens}.

A separate line of work obtains timing from a pre-generated modality. Audio-conditioned video methods synchronize visual dynamics to an existing soundtrack~\cite{weng2025mtv,lee2025syncphony,song2026syncdit}, while video-to-audio systems synthesize sound aligned with an input video~\cite{luo2023difffoley,cheng2025mmaudio,kushwaha2025vintage,chen2025multifoley}. Talking-face and avatar systems likewise animate visual content from pre-generated speech~\cite{prajwal2020wav2lip,zhang2023sadtalker,shen2023difftalk,yu2023diffusionprior,aneja2024facetalk,peng2024synctalk,guan2025audcast,corona2025vlogger,cui2025hallo2}. These methods achieve temporal alignment by treating audio or video as an external timing anchor.

Existing approaches therefore either control timing within a single generated modality or inherit it from an externally provided modality. TimeSteer instead relocates model-generated speech together with its visual articulation inside a pretrained joint audio-visual diffusion model. It performs this scheduling entirely at inference time from user-specified intervals, without external speech or additional training.

\section{Method and Benchmark}
\subsection{Task Formulation}

Let $c$ denote a text prompt describing a scene and containing an ordered collection $\mathcal{U}$ of $N$ quoted utterances. For each utterance indexed by $u\in\{1,\ldots,N\}$, the user specifies a target interval $\mathcal{T}^u=[t_s^u,t_e^u]$, where $t_s^u$ and $t_e^u$ denote its target start and end times in the generated video, respectively. Given a pretrained joint audio-visual diffusion model $\mathcal{G}_{\theta}$ with frozen parameters $\theta$, our goal is to construct a training-free scheduling operator $\mathcal{S}$ for inference-time speech scheduling:
\begin{equation}
(A,V)
=
\mathcal S(\mathcal G_\theta,c,\{\mathcal T^u\}_{u=1}^{N}),
\end{equation}
where $A$ and $V$ denote the generated audio and video, respectively. The generated output should align the realized interval of every utterance $u$ with $\mathcal{T}^u$ while preserving the generation quality of the pretrained model.

\subsection{TimeSteer}

\begin{figure*}[ht]
    \centering
    \includegraphics[width=\linewidth]{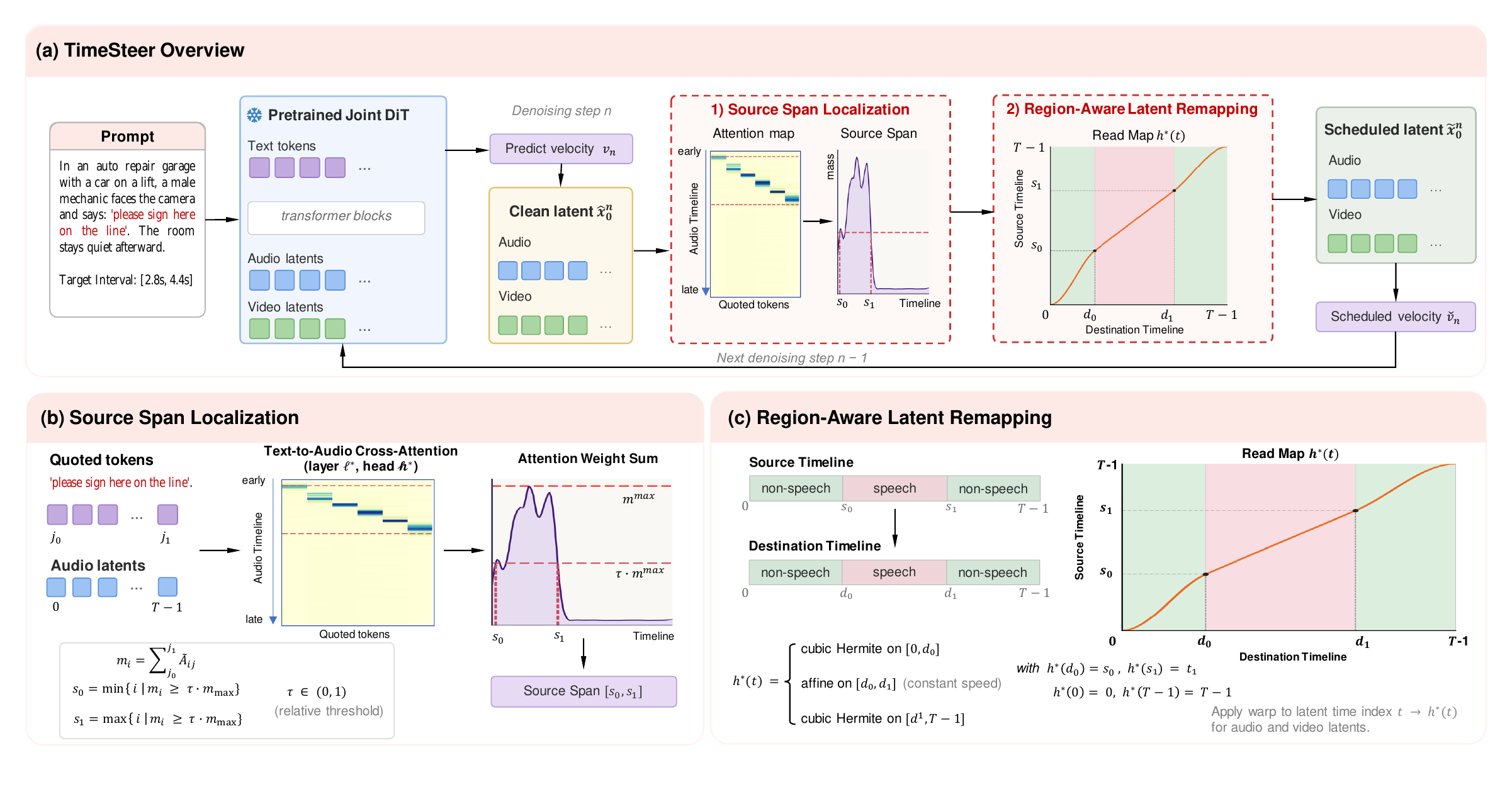}
    \caption{\textbf{Overview of TimeSteer}. (a) Given target intervals, TimeSteer localizes each utterance's source span and relocates its predicted audio-visual content. (b) Source Span Localization estimates source spans from text-to-audio cross-attention. (c) Region-Aware Latent Remapping relocates the content through its read map.}
    \label{fig:methods}
\end{figure*}

TimeSteer is a training-free framework for inference-time speech scheduling in joint audio-visual diffusion models. As illustrated in Fig.~\ref{fig:methods}, it enables precise temporal control by intervening on the predicted clean latent throughout the denoising process, without modifying the pretrained model parameters.

At each denoising step $n$, the diffusion model predicts the clean latent $\hat{x}_0^n$ from the current noisy latent following the standard flow-matching formulation~\cite{lipman2023flowmatching,liu2023rectifiedflow}, i.e., $\hat{x}_0^n=x_n-\sigma_n v_n$, where $x_n$, $v_n$, and $\sigma_n$ denote the noisy latent, predicted velocity, and noise level, respectively. Before the prediction $\hat{x}_0^n$ is used for the next denoising update, the scheduling operator acts on the predicted clean latent as
\begin{equation}
\tilde{x}_0^n
=
\mathcal{S}\!\left(\hat{x}_0^n\right),
\label{eq:x0-steer}
\end{equation}
where $\tilde{x}_0^n$ denotes the scheduled clean latent. The scheduled clean latent is then converted back into the corresponding velocity prediction following the standard flow-matching update, allowing the remaining denoising process to proceed unchanged.

TimeSteer implements the scheduling operator $\mathcal{S}$ in two successive stages. First, \textbf{Source Span Localization} identifies the source span of each quoted utterance from the cross-attention map. Then, using these localized spans, \textbf{Region-Aware Latent Remapping} constructs a read map from the target latent timeline to the source latent timeline. The map specifies which source coordinate is sampled at each target coordinate, thereby relocating each utterance to its target interval while preserving temporal continuity. The procedure is provided in the supplementary material.

\subsubsection{Source Span Localization}
\label{sec:span-localization}

Source Span Localization builds on our observation that a timing-sensitive text-to-audio cross-attention head in the frozen DiT produces a temporally localized response for each utterance. This response traces the utterance’s model-implied source span along the audio latent timeline. Based on that, we can get an explicit span estimate before decoding, as illustrated in Fig.~\ref{fig:methods}(b).

Let $T$ denote the latent temporal length and $\mathcal{A}_0^n$ the audio branch of the predicted clean latent $\hat{x}_0^n$, which is the model's estimate of the final clean latent at denoising step $n$. We replay $\mathcal{A}_0^n$ through the frozen DiT and extract a text-to-audio cross-attention map $\tilde{A}\in\mathbb{R}^{T\times L_c}$ from the timing-sensitive attention layer $\mathscr{l}*$ with head $\mathscr{h}^*$, where $L_c$ denotes the number of text tokens. Each entry $\tilde{A}_{ij}$ measures the correspondence between audio latent position $i$ and text token $j$.

For an utterance $u$ spanning the quoted token range $[j_0^u,j_1^u]$, where $j_0^u$ and $j_1^u$ are its first and last token indices, we aggregate the attention weights over its quoted tokens:
\begin{equation}
m_i^u
=
\sum_{j=j_0^u}^{j_1^u}
\tilde{A}_{ij},
\qquad
i=0,\ldots,T-1.
\label{eq:row-mass}
\end{equation}
The resulting scores measure the correspondence between each audio latent position and utterance $u$.

Let $m_{\max}^u=\max_i m_i^u$. Using a relative threshold $\tau\in(0,1)$, we estimate the source span as
\begin{equation}
\begin{aligned}
S^u
&=
[s_0^u,s_1^u],\\
s_0^u
&=
\min\left\{i\,\middle|\,m_i^u\ge\tau m_{\max}^u\right\},\\
s_1^u
&=
\max\left\{i\,\middle|\,m_i^u\ge\tau m_{\max}^u\right\}.
\end{aligned}
\label{eq:source-span}
\end{equation}
The resulting source span $S^u$ is subsequently used by Region-Aware Latent Remapping.

\subsubsection{Region-Aware Latent Remapping}
\label{sec:hybrid-remapping}

Given the localized source spans and target intervals, the remaining task is to relocate the predicted audio-visual content from each source span to its target interval while preserving temporal continuity, as illustrated in Fig.~\ref{fig:methods}(c).

Our key idea is to relocate the predicted clean latent content without regenerating it. We construct a read map over the latent timeline that specifies which source position in the predicted clean latent is sampled at each destination position.

For each utterance $u$, mapping the target interval $\mathcal{T}^u$ onto the latent timeline yields the destination speech interval $[d_0^u,d_1^u]$, which is paired with the localized source span $S^u=[s_0^u,s_1^u]$. The source and destination intervals need not have the same duration.

We define a destination-to-source read map
\begin{equation}
h:[0,T-1]\rightarrow[0,T-1],
\label{eq:read-map}
\end{equation}
where $h(t)$ specifies the source coordinate sampled at destination coordinate $t$. Because the source and destination speech intervals may differ in duration, we align their endpoints by imposing $h(d_0^u)=s_0^u$ and $h(d_1^u)=s_1^u$. We also set $h(0)=0$ and $h(T-1)=T-1$ to keep the clip boundaries fixed. In implementation, we evaluate the continuous map $h$ at discrete destination indices. When $h(t)$ falls between two source indices, the corresponding latent is obtained by linearly interpolating the two neighboring latent vectors, with weights determined by the fractional part of $h(t)$.

The derivative $h'(t)=dh(t)/dt$ is the local slope of the read map. Because $h$ maps destination time to source time, this slope scales the original speaking rate within each speech interval and can therefore be interpreted as the local speaking-rate ratio, i.e., the output speaking rate divided by the original speaking rate. The same temporal scaling applies to the associated visual articulation.

Speech intervals and non-speech gaps require different remapping strategies.
Specifically, we construct $h$ using two complementary components: \textbf{Minimal-Distortion Remapping}, which applies an affine map within each speech interval, and \textbf{Curvature Bridging}, which smoothly interpolates the read map across non-speech gaps.

\paragraph{Minimal-Distortion Remapping for Speech Intervals.}
Within each speech interval, we aim to preserve the original speaking rate, corresponding to $h'(t)=1$. Under the source--destination endpoint constraints, we therefore minimize the accumulated deviation from this value:
\begin{equation}
\begin{aligned}
h_u^\star
&=
\arg\min_h\quad
\int_{d_0^u}^{d_1^u}
\bigl(h'(t)-1\bigr)^2\,dt,
\\
&\text{s.t.}\quad
h(d_0^u)=s_0^u,
\qquad
h(d_1^u)=s_1^u.
\end{aligned}
\label{eq:speech-variational}
\end{equation}
The resulting minimizer is the affine map
\begin{equation}
h_u^\star(t)
=
s_0^u
+
\kappa_u
\bigl(t-d_0^u\bigr),
\qquad
t\in[d_0^u,d_1^u],
\label{eq:minimal-distortion-map}
\end{equation}
where
$\kappa_u=\frac{s_1^u-s_0^u}{d_1^u-d_0^u}$
is its constant slope and thus the speaking-rate ratio for utterance $u$. Thus, each speech interval uses a simple linear read map.

\paragraph{Curvature Bridging for Non-speech Gaps.}

Across a non-speech gap, the slopes inherited from neighboring speech intervals may differ, and we seek to connect them with as little temporal-rate variation as possible. Since $h''(t)=dh'(t)/dt$ measures the change in the read-map slope, $h''(t)=0$ represents the ideal case of no rate variation. We therefore minimize its accumulated squared magnitude under the boundary constraints.

Consider the interior gap between utterances $u$ and $u+1$, bounded by $(d_1^u,s_1^u)$ and $(d_0^{u+1},s_0^{u+1})$. Its boundary slopes are inherited directly from the adjacent affine speech maps as $\kappa_u$ and $\kappa_{u+1}$. The resulting variational problem is:
\begin{equation}
\begin{aligned}
h_{u,u+1}^\star
&=
\arg\min_h\quad
\int_{d_1^u}^{d_0^{u+1}}
\bigl(h''(t)\bigr)^2\,dt,
\\
&\text{s.t.}\quad
h(d_1^u)=s_1^u,
\quad
 h(d_0^{u+1})=s_0^{u+1},
\\
&\phantom{\text{s.t.}\quad}
h'(d_1^u)=\kappa_u,
\quad
h'(d_0^{u+1})=\kappa_{u+1}.
\end{aligned}
\label{eq:gap-variational}
\end{equation}
This optimization yields the unique cubic solution
\begin{equation}
\begin{aligned}
h_{u,u+1}^\star(t)
&=
\sum_{r=0}^{3}
a_{u,r}\bigl(t-d_1^u\bigr)^r,
\\
&\quad t\in[d_1^u,d_0^{u+1}],
\end{aligned}
\label{eq:gap-solution}
\end{equation}
where the coefficients $\{a_{u,r}\}_{r=0}^{3}$ are uniquely determined by the four boundary constraints in Eq.~\eqref{eq:gap-variational}. Thus, each interior gap uses a cubic read map that smoothly connects the adjacent linear speech maps. 

\subsection{SpeechShift Benchmark}
\label{sec:benchmark}
SpeechShift is a benchmark for interval-level speech scheduling in text-to-audio-visual generation. Each sample follows the task representation $(c,\mathcal{U},\{\mathcal{T}^u\}_{u=1}^{N})$ and contains a scene prompt, an ordered set of quoted utterances, their speaker roles and target intervals, and metadata describing the scene and challenge conditions. 

In detail, the benchmark contains 400 prompts across 102 scenes and 600 utterance-level targets. It is balanced across four speaking structures defined by speaker and utterance counts: \emph{Single-Speaker Single-Utterance (SS--SU)}, \emph{Single-Speaker Multi-Utterance (SS--MU)}, \emph{Multi-Speaker Single-Utterance (MS--SU)}, and \emph{Multi-Speaker Multi-Utterance (MS--MU)}, with 100 prompts per structure. Within each structure, the prompts further form four condition groups by combining clean or interfering acoustics with static or challenging visual actions. The acoustic conditions include ambient noise, transient sounds, competing speech, and mixed interference, while the visual conditions include motion, occlusion, and action-coupled speaking.
Target intervals are assigned solely from the text annotations, and the durations are adjusted according to utterance length to cover both fast and slow speech. For two-utterance cases, we use non-overlapping target intervals with gaps ranging from short to long.

\subsubsection{Evaluation Protocol}
\label{sec:benchmark-metrics}

SpeechShift evaluates two complementary objectives: placing each utterance within its target interval and retaining the overall quality of the generated audio-visual content. We assess them through interval controllability and generation quality, respectively.

\paragraph{Interval controllability.}

Let $\mathcal{E}$ denote the set of evaluated utterance instances. For each $u\in\mathcal{E}$, we transcribe and word-align the generated clip to estimate the realized interval $\hat{\mathcal{T}}^u=[\hat{t}_s^u,\hat{t}_e^u]$ corresponding to the target interval $\mathcal{T}^u=[t_s^u,t_e^u]$. We define the absolute onset and offset errors as $e_s^u=|\hat{t}_s^u-t_s^u|$ and $e_e^u=|\hat{t}_e^u-t_e^u|$, respectively, and report their means over $\mathcal{E}$ as $\mathrm{mE}_s$ and $\mathrm{mE}_e$. The temporal Intersection-over-Union (IoU) is
\begin{equation}
\mathrm{IoU}^u
=
\frac{
\left|\hat{\mathcal{T}}^u\cap\mathcal{T}^u\right|
}{
\left|\hat{\mathcal{T}}^u\cup\mathcal{T}^u\right|
},
\end{equation}
and the reported IoU is its mean over $u\in\mathcal{E}$. We further report the hit rate
\begin{equation}
\mathrm{HR}_{\delta}
=
\frac{1}{|\mathcal{E}|}
\sum_{u\in\mathcal{E}}
\mathbb{I}
\left(
\max(e_s^u,e_e^u)
\leq
\delta
\right),
\end{equation}
which measures the proportion of utterances whose two boundary errors both fall within the tolerance $\delta$.

\paragraph{Generation quality.}

Alongside interval controllability, we report one representative metric for each quality dimension: WER for speech intelligibility, Production Quality (PQ) for audio naturalness~\cite{tjandra2025audiobox}, MUSIQ for no-reference visual quality~\cite{ke2021musiq}, and LSE-C for audio-visual synchronization within each target interval~\cite{chung2017syncnet}.

\section{Experiments}
\label{sec:experiments}

\subsection{Experimental Settings}
\label{sec:exp-setups}

We evaluate TimeSteer on SpeechShift using two representative joint audio-visual diffusion backbones, LTX-2~\cite{hacohen2025ltx2} and daVinci-MagiHuman~\cite{chern2026speed}, which adopt different cross-modal architectures, comparing against four representative training-free baselines: (\textbf{i}) \emph{Uncontrolled}, standard sampling without temporal intervention; (\textbf{ii}) \emph{Textual Timing}, which adds explicit timing instructions to the prompt~\cite{xu2024picoaudio,huang2025tempocontrol}; (\textbf{iii}) \emph{FreeAudio}~\cite{zhang2025freeaudio}, which binds each utterance to its target attention window through constrained attention aggregation; and (\textbf{iv}) \emph{Prompt Relay}~\cite{chen2026promptrelay}, which concentrates each prompt segment in its target region through a distance-dependent attention penalty.
All methods use the same backbone, prompts, target intervals, and inference settings. Unless otherwise stated, we generate 5-second audio-visual clips with seed 42. The implementation details are provided in the supplementary material.

\subsection{Main Results}

\begin{table*}[ht]
\centering
\footnotesize
\setlength{\tabcolsep}{5pt}
\resizebox{\textwidth}{!}{%
\begin{tabular}{lcccc|c|cc|c}
\toprule
&
\multicolumn{4}{c|}{\textbf{Interval Controllability}}
&
\multicolumn{1}{c|}{\textbf{Semantic Fidelity}}
&
\multicolumn{2}{c|}{\textbf{Perceptual Quality}}
&
\textbf{Cross-modal}\\

\cmidrule(lr){2-5}
\cmidrule(lr){6-6}
\cmidrule(lr){7-8}
\cmidrule(l){9-9}

\textbf{Method}
&
\textbf{HR$_{0.2}$}$\uparrow$
&
\textbf{IoU}$\uparrow$
&
\textbf{$mE_s$ (s)}$\downarrow$
&
\textbf{$mE_e$ (s)}$\downarrow$
&
\textbf{WER}$\downarrow$
&
\textbf{PQ}$\uparrow$
&
\textbf{MUSIQ}$\uparrow$
&
\textbf{LSE-C}$\uparrow$
\\

\midrule

\multicolumn{9}{l}{\textbf{LTX-2}}\\

Uncontrolled & 0.21 & 0.63 & 0.56 & 0.54 & \textbf{0.05} & 5.36 & 52.26 & 3.13 \\
Textual Timing & 0.19 & 0.60 & 0.59 & 0.59 & 0.07 & 5.36 & 52.26 & \textbf{3.18} \\
FreeAudio & 0.19 & 0.54 & 0.49 & 0.67 & 0.35 & \textbf{5.92} & \textbf{52.55} & 2.92 \\
Prompt Relay & 0.31 & 0.68 & 0.27 & 0.35 & 0.20 & 5.52 & 52.00 & 3.06 \\

\rowcolor{gray!10}
TimeSteer (Ours) & \textbf{0.73} & \textbf{0.87} & \textbf{0.11} & \textbf{0.15} & 0.07 & 5.32 & 52.51 & 3.10 \\

\addlinespace

\multicolumn{9}{l}{\textbf{daVinci-MagiHuman}}\\

Uncontrolled & 0.09 & 0.63 & 0.63 & 0.49 & \textbf{0.06} & 5.53 & 69.36 & 3.67 \\
Textual Timing & 0.08 & 0.61 & 0.64 & 0.53 & 0.09 & 5.50 & 68.98 & \textbf{3.87} \\
FreeAudio & 0.01 & 0.16 & 1.04 & 1.40 & 0.95 & 5.24 & 66.80 & 1.63 \\
Prompt Relay & 0.21 & 0.68 & 0.39 & 0.32 & 0.29 & \textbf{5.61} & \textbf{70.87} & 3.66 \\

\rowcolor{gray!10}
TimeSteer (Ours) & \textbf{0.53} & \textbf{0.79} & \textbf{0.19} & \textbf{0.18} & 0.08 & 5.42 & 69.08 & 3.56 \\

\bottomrule
\end{tabular}
}
\caption{
\textbf{{Main results.}} SpeechShift comparison with training-free baselines on two joint audio-visual diffusion backbones. Complete quality results appear in the supplementary material. Best results are in bold. $\uparrow$/$\downarrow$ indicates higher/lower is better.
}
\label{tab:quality}
\end{table*}

Table~\ref{tab:quality} shows that TimeSteer substantially improves interval controllability over Uncontrolled sampling on both backbones. On LTX-2, HR$_{0.2}$ increases from $0.21$ to $0.73$ and IoU from $0.63$ to $0.87$; on daVinci-MagiHuman, HR$_{0.2}$ increases from $0.09$ to $0.53$ and IoU from $0.63$ to $0.79$. TimeSteer also achieves the lowest onset and offset errors on both architectures.
Compared with the other control methods, Textual Timing and FreeAudio provide limited or inconsistent improvements, while Prompt Relay achieves better control but remains behind TimeSteer across all four controllability metrics. Meanwhile, TimeSteer maintains generation-quality metrics comparable to uncontrolled sampling. Overall, the results demonstrate strong and consistent interval controllability across both architectures without a substantial loss in generation quality.
\subsection{Ablation Study}
\label{sec:ablations}

We conduct three ablation studies to examine source-span localization, the choice of remapping space, and the region-aware read-map design. Each study is evaluated on 50 randomly sampled cases from SpeechShift.

\subsubsection{Effect of Source Span Localization}
\label{sec:ablation-localization}

We examine four ways to localize the model-implied source span during denoising. At each step $n$, the current latent $x_n$ remains corrupted by noise, while $\hat{x}_0^n$ is the model's estimate of the final clean latent. \textbf{Attn@$\hat{x}_0^n$} feeds this clean estimate back into the transformer and localizes the quoted utterance from its text-to-audio attention weights, which is the strategy used by TimeSteer. \textbf{Attn@$x_n$} instead reads the attention produced by the standard forward pass on the noisy latent. As alternatives that do not use attention, \textbf{Decode@$\hat{x}_0^n$} and \textbf{Decode@$x_n$} decode the corresponding audio latent into an intermediate waveform and estimate the source span from its energy-based speech activity. 

\begin{figure}[ht]
\centering
\includegraphics[width=\linewidth]{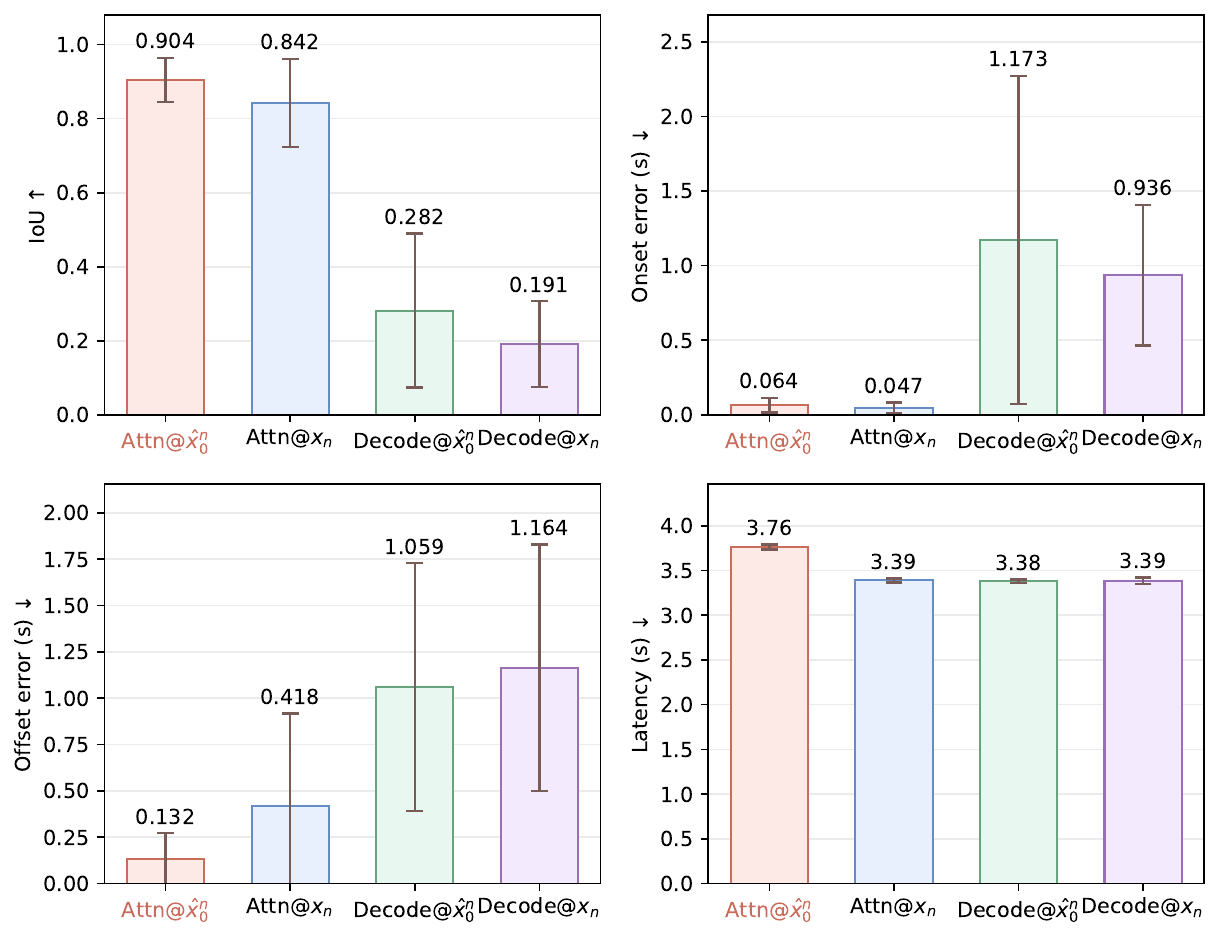}
\caption{\textbf{Source-span localization ablation.} Mean performance of four localization strategies; error bars show standard deviations across the evaluation subset and steps.}
\label{fig:ablation-localization}
\end{figure}

As shown in Fig.~\ref{fig:ablation-localization}, Attn@$\hat{x}_0^n$ achieves the highest interval IoU and the lowest onset and offset errors. Attn@$x_n$ retains comparable onset accuracy but produces less reliable offsets, showing that the clean-latent estimate exposes sharper utterance boundaries. Both decode-based variants perform substantially worse because speech activity is not yet stably formed in intermediate waveforms. Attn@$\hat{x}_0^n$ therefore provides the most reliable source span before decoding, with only a modest latency overhead, and is used throughout TimeSteer.

\subsubsection{Effect of Remapping Space}
\label{sec:ablation-x0-traj}

We compare four intervention spaces to determine where temporal remapping should be applied during generation. \textbf{Remap@$\hat{x}_0^n$}, the strategy used by TimeSteer, applies the read map to the predicted clean audio-visual latent and converts the remapped estimate back into the update required by the original sampler. \textbf{Remap@$v_n$} applies the same map directly to the predicted velocity before the sampler update, whereas \textbf{Remap@$x_n$} remaps the updated noisy latent passed to the next denoising step. \textbf{Post-hoc} performs no intervention during denoising and instead remaps the final audio-visual latent once before decoding.

\begin{figure}[t]
\centering
\includegraphics[width=\linewidth]{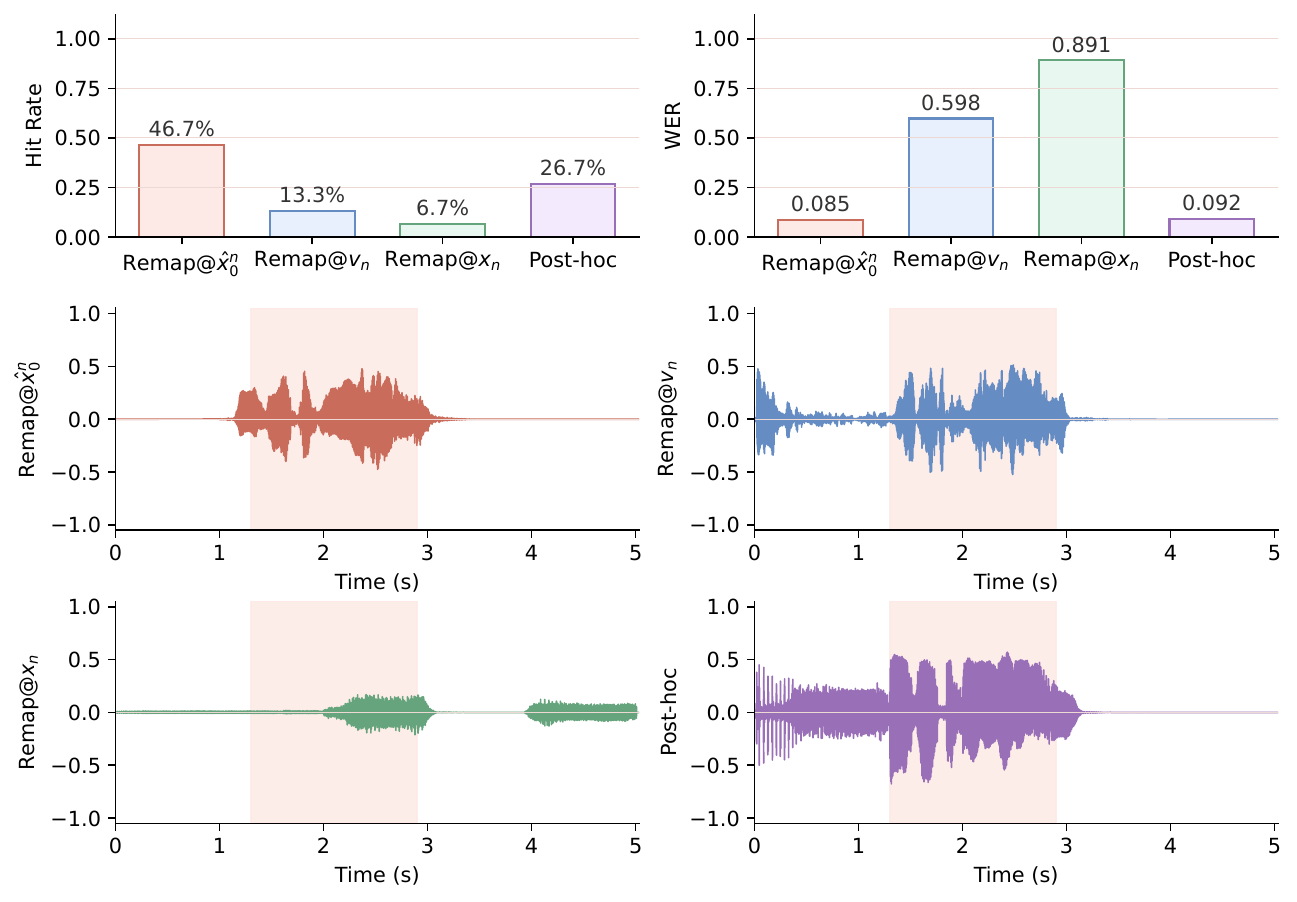}
\caption{\textbf{Remapping-space ablation.} Top: $\mathrm{HR}_{0.2}$ and WER for different intervention spaces. Bottom: representative decoded waveforms; shading marks target intervals.}
\label{fig:ablation-x0-traj}
\end{figure}

As shown in Fig.~\ref{fig:ablation-x0-traj}, Remap@$\hat{x}_0^n$ achieves the highest $\mathrm{HR}_{0.2}$ and the lowest WER. The representative waveforms provide complementary qualitative evidence: Remap@$\hat{x}_0^n$ produces well-formed speech activity within the shaded target intervals while maintaining a comparatively clean background, whereas the other three strategies all exhibit varying degrees of noisy residuals. Directly remapping $v_n$ or $x_n$ further disrupts the denoising trajectory, producing incomplete or nearly silent utterances and substantially degrading transcription accuracy. Post-hoc remapping largely preserves the generated content but cannot reliably relocate it, because no subsequent denoising steps remain to consolidate the new temporal arrangement or suppress the introduced artifacts. In contrast, remapping $\hat{x}_0^n$ presents the sampler with a rescheduled clean-latent prediction at every step, allowing the relocated speech to be progressively refined within its target interval.

\subsubsection{Effect of Read-Map Design}
\label{sec:ablation-hybrid-geom}

The region-aware read map matches each region with the geometry suited to its temporal role: a linear map applies a uniform rate adjustment within each utterance, preserving its internal temporal structure, while cubic bridges smoothly accommodate the necessary rate transitions across non-speech gaps. To test this region-aware design, we compare it with two alternatives that apply one rule throughout the timeline. \textbf{Piecewise-linear} connects every consecutive source--destination boundary pair with a line segment. It therefore maintains a constant slope within each segment but permits abrupt slope changes at speech boundaries. \textbf{Global PCHIP}~\cite{fritsch1980monotone} instead constructs a shape-preserving monotone cubic Hermite interpolant through all boundary pairs. It maintains slope continuity across the full timeline but does not constrain the slope to remain constant within speech intervals.

As shown in Table~\ref{tab:ablation-hybrid-geom}, our region-aware design consistently outperforms both alternatives in scheduling accuracy. It achieves a Span Win of 67.9\%, more than three times the 19.4\% obtained by Global PCHIP, while also attaining the highest $\mathrm{HR}_{0.2}$ and temporal IoU. This consistent per-span advantage supports assigning distinct geometries to speech intervals and non-speech gaps. 

\begin{table}[t]
\centering
\footnotesize
\setlength{\tabcolsep}{4pt}
\begin{tabular}{lcccc}
\toprule
\textbf{Read-map design} & $\textbf{HR}_{0.2}$$\uparrow$ & \textbf{WER}$\downarrow$ & \textbf{IoU}$\uparrow$ & \textbf{Span Win}$\uparrow$ \\
\midrule
Piecewise-linear & 48.6\% & \textbf{7.88\%} & 69.8\% & 12.7\% \\
Global PCHIP & 52.3\% & 8.92\% & 75.8\% & 19.4\% \\
\rowcolor[gray]{0.95} Region-Aware (Ours) & \textbf{69.8\%} & 8.05\% & \textbf{84.3\%} & \textbf{67.9\%} \\
\bottomrule
\end{tabular}
\caption{\textbf{Read-map design ablation}. IoU measures temporal overlap, while Span Win indicates how often each geometry aligns best. All values are percentages; best results are bold.}
\label{tab:ablation-hybrid-geom}
\end{table}

\section{Conclusion}

Joint audio-visual diffusion models can synthesize coupled speech and visual articulation, yet their temporal organization remains an implicit model decision. We show that such timing can be exposed and redirected during generation. Specifically, we uncover that a timing-sensitive cross-attention head reveals each utterance's source span, while the predicted clean latent contains the coupled content required for temporal relocation. Building on these discoveries, TimeSteer localizes each source span and transfers its content to any user-specified target interval during denoising through a region-aware read map, without retraining the generator. We further introduce SpeechShift, the first benchmark for interval-level speech scheduling in joint audio-visual generation, covering diverse speaking structures. Across LTX-2 and daVinci-MagiHuman, TimeSteer substantially improves interval controllability over training-free baselines while retaining competitive generation quality, and our ablations validate its key localization and remapping designs. 
Together, TimeSteer and SpeechShift establish a practical method and evaluation foundation for inference-time speech scheduling. We hope our work will inform future research on temporal control in audio-visual generation.

\clearpage
\bibliography{aaai2027}

@article{polyak2024moviegen,
  title   = {Movie Gen: A Cast of Media Foundation Models},
  author  = {Polyak, Adam and Zohar, Amit and Brown, Andrew and Tjandra, Andros and Sinha, Animesh and Lee, Ann and Vyas, Apoorv and Shi, Bowen and Ma, Chih-Yao and Chuang, Ching-Yao and others},
  journal = {arXiv preprint arXiv:2410.13720},
  year    = {2024}
}

@inproceedings{xing2024seeing,
  title     = {Seeing and Hearing: Open-Domain Visual-Audio Generation with Diffusion Latent Aligners},
  author    = {Xing, Yazhou and He, Yingqing and Tian, Zeyue and Wang, Xintao and Chen, Qifeng},
  booktitle = {Proceedings of the IEEE/CVF Conference on Computer Vision and Pattern Recognition},
  pages     = {7151--7161},
  year      = {2024}
}

@inproceedings{wang2025animate,
  title     = {Animate and Sound an Image},
  author    = {Wang, Xihua and Song, Ruihua and Li, Chongxuan and Cheng, Xin and Li, Boyuan and Wu, Yihan and Wang, Yuyue and Xu, Hongteng and Wang, Yunfeng},
  booktitle = {Proceedings of the IEEE/CVF Conference on Computer Vision and Pattern Recognition},
  pages     = {23369--23378},
  year      = {2025}
}

@article{hacohen2025ltx2,
  title   = {{LTX-2}: Efficient Joint Audio-Visual Foundation Model},
  author  = {HaCohen, Yoav and Brazowski, Benny and Chiprut, Nisan and Bitterman, Yaki and Kvochko, Andrew and Berkowitz, Avishai and Shalem, Daniel and Lifschitz, Daphna and Moshe, Dudu and Porat, Eitan and others},
  journal = {arXiv preprint arXiv:2601.03233},
  year    = {2025}
}

@article{nava2026,
  title   = {{NAVA}: Native Audio-Visual Alignment for Generation},
  author  = {Baidu ERNIE Team},
  journal = {arXiv preprint arXiv:2605.30073},
  year    = {2026}
}

@article{wang2025javisdit,
  title   = {{JavisDiT}: Joint Audio-Video Diffusion Transformer with Hierarchical Spatio-Temporal Prior Synchronization},
  author  = {Wang, Yan and Chen, Yixin and Niu, Di and Zhang, Ya and Wang, Hui and Chen, Siheng and others},
  journal = {arXiv preprint arXiv:2503.23377},
  year    = {2025}
}

@article{li2025harmony,
  title   = {Harmony: Harmonizing Audio and Video Generation through Cross-Task Synergy},
  author  = {Li, Zeyu and others},
  journal = {arXiv preprint arXiv:2511.21579},
  year    = {2025}
}

@article{wang2024syncflow,
  title   = {{SyncFlow}: Synchronized Joint Audio-Video Generation with Dual Diffusion Transformers},
  author  = {Wang, Yan and others},
  journal = {arXiv preprint arXiv:2412.15220},
  year    = {2024}
}

@article{chen2025unison,
  title   = {Unison: Harmonizing Motion, Speech, and Sound for Human-Centric Audio-Video Generation},
  author  = {Chen, Zhaoxin and others},
  journal = {arXiv preprint arXiv:2605.08729},
  year    = {2025}
}

@inproceedings{ruan2023mmdiffusion,
  title     = {{MM-Diffusion}: Learning Multi-Modal Diffusion Models for Joint Audio and Video Generation},
  author    = {Ruan, Yang and Liu, Junwei and others},
  booktitle = {Proceedings of the IEEE/CVF Conference on Computer Vision and Pattern Recognition},
  year      = {2023}
}

@article{zhang2026mmcontrol,
  title   = {{MMControl}: Unified Multi-Modal Control for Joint Audio-Video Generation},
  author  = {Zhang, Yingqing and others},
  journal = {arXiv preprint arXiv:2604.19679},
  year    = {2026}
}

@article{ma2026itsjav,
  title   = {Inference-Time Scaling for Joint Audio-Video Generation},
  author  = {Ma, Yuhang and others},
  journal = {arXiv preprint arXiv:2606.03183},
  year    = {2026}
}

@article{wang2026mova,
  title   = {{MOVA}: Towards Scalable and Synchronized Video-Audio Generation},
  author  = {Wang, Zhen and others},
  journal = {arXiv preprint arXiv:2602.08794},
  year    = {2026}
}

@article{low2025ovi,
  title   = {Ovi: Twin Backbone Cross-Modal Fusion for Audio-Video Generation},
  author  = {Low, Chetwin and Wang, Weimin and Katyal, Calder},
  journal = {arXiv preprint arXiv:2510.01284},
  year    = {2025}
}

@article{liu2026avgenbench,
  title   = {{AVGen-Bench}: A Task-Driven Benchmark for Multi-Granular Evaluation of Text-to-Audio-Video Generation},
  author  = {Liu, Ziwei and others},
  journal = {arXiv preprint arXiv:2604.08540},
  year    = {2026}
}

@inproceedings{kreuk2023audiogen,
  title     = {{AudioGen}: Textually Guided Audio Generation},
  author    = {Kreuk, Felix and Synnaeve, Gabriel and Polyak, Adam and Singer, Uri and D{\'e}fossez, Alexandre and Copet, Jade and Parikh, Devi and Taigman, Yaniv and Adi, Yossi},
  booktitle = {International Conference on Learning Representations},
  year      = {2023}
}

@inproceedings{liu2023audioldm,
  title     = {{AudioLDM}: Text-to-Audio Generation with Latent Diffusion Models},
  author    = {Liu, Haohe and Chen, Zehua and Yuan, Yi and Mei, Xinhao and Liu, Xubo and Mandic, Danilo and Wang, Wenwu and Plumbley, Mark D.},
  booktitle = {Proceedings of the International Conference on Machine Learning},
  year      = {2023}
}

@article{huang2024makeanaudio2,
  title   = {Make-An-Audio 2: Temporal-Enhanced Text-to-Audio Generation},
  author  = {Huang, Rongjie and others},
  journal = {arXiv preprint arXiv:2305.18474},
  year    = {2024}
}

@article{huang2024audiocomposer,
  title   = {{AudioComposer}: Towards Fine-Grained Text-to-Audio Generation},
  author  = {Huang, Rongjie and others},
  journal = {arXiv preprint arXiv:2402.10012},
  year    = {2024}
}

@article{xu2024picoaudio,
  title   = {{PicoAudio}: Enabling Precise Timestamp and Frequency Controllability of Audio Events in Text-to-Audio Generation},
  author  = {Xu, Hao and others},
  journal = {arXiv preprint arXiv:2407.02869},
  year    = {2024}
}

@article{li2025picoaudio2,
  title   = {{PicoAudio2}: Temporally Strong Text-to-Audio Generation with Context and Timestamp Matrix},
  author  = {Li, Yiming and others},
  journal = {arXiv preprint arXiv:2509.00683},
  year    = {2025}
}

@article{wang2025controlaudio,
  title   = {{ControlAudio}: Tackling Text-Guided, Timing-Indicated and Intelligible Audio Generation via Progressive Diffusion Modeling},
  author  = {Wang, Yixin and others},
  journal = {arXiv preprint arXiv:2510.08878},
  year    = {2025}
}

@article{zhang2025freeaudio,
  title   = {{FreeAudio}: Training-Free Timing Planning for Controllable Long-Form Text-to-Audio Generation},
  author  = {Zhang, Yu and others},
  journal = {arXiv preprint arXiv:2507.08557},
  year    = {2025}
}

@inproceedings{luo2023difffoley,
  title     = {{Diff-Foley}: Synchronized Video-to-Audio Synthesis with Latent Diffusion Models},
  author    = {Luo, Simian and Yan, Chuanhao and Hu, Chenxu and Zhao, Hang},
  booktitle = {Advances in Neural Information Processing Systems},
  volume    = {36},
  year      = {2023}
}

@inproceedings{cheng2025mmaudio,
  title     = {{MMAudio}: Taming Multimodal Joint Training for High-Quality Video-to-Audio Synthesis},
  author    = {Cheng, Ho Kei and Ishii, Masato and Hayakawa, Akio and Shibuya, Takashi and Schwing, Alexander and Mitsufuji, Yuki},
  booktitle = {Proceedings of the IEEE/CVF Conference on Computer Vision and Pattern Recognition},
  pages     = {28901--28911},
  year      = {2025}
}

@inproceedings{kushwaha2025vintage,
  title     = {{VinTAGe}: Joint Video and Text Conditioning for Holistic Audio Generation},
  author    = {Kushwaha, Saksham Singh and Tian, Yapeng},
  booktitle = {Proceedings of the IEEE/CVF Conference on Computer Vision and Pattern Recognition},
  pages     = {13529--13539},
  year      = {2025}
}

@inproceedings{chen2025multifoley,
  title     = {Video-Guided Foley Sound Generation with Multimodal Controls},
  author    = {Chen, Ziyang and Seetharaman, Prem and Russell, Bryan and Nieto, Oriol and Bourgin, David and Owens, Andrew and Salamon, Justin},
  booktitle = {Proceedings of the IEEE/CVF Conference on Computer Vision and Pattern Recognition},
  pages     = {18770--18781},
  year      = {2025}
}

@article{huang2025tempocontrol,
  title   = {{TempoControl}: Temporal Attention Guidance for Text-to-Video Models},
  author  = {Huang, Yuchao and others},
  journal = {arXiv preprint arXiv:2510.02226},
  year    = {2025}
}

@article{chen2026promptrelay,
  title   = {{Prompt Relay}: Inference-Time Temporal Control for Multi-Event Video Generation},
  author  = {Chen, Gordon and Huang, Ziqi and Liu, Ziwei},
  journal = {arXiv preprint arXiv:2604.10030},
  year    = {2026}
}

@article{weng2025mtv,
  title   = {Audio-Sync Video Generation with Multi-Stream Temporal Control},
  author  = {Weng, Shuchen and Zheng, Haojie and Chang, Zheng and Li, Si and Shi, Boxin and Wang, Xinlong},
  journal = {arXiv preprint arXiv:2506.08003},
  year    = {2025}
}

@article{lee2025syncphony,
  title   = {{Syncphony}: Synchronized Audio-to-Video Generation with Diffusion Transformers},
  author  = {Lee, Jibin and others},
  journal = {arXiv preprint arXiv:2509.21893},
  year    = {2025}
}

@article{song2026syncdit,
  title   = {{SyncDIT}: Audio-Visual Aligned Video Generation with Audio Synchronization Feature},
  author  = {Song, Quanyue and Guo, Zhizhi and He, Yishan and Wang, Zhihao and He, Zhixiang and Zhang, Chi and Jiang, Caigui and Li, Xuelong},
  journal = {Vicinagearth},
  volume  = {3},
  pages   = {5},
  year    = {2026},
  doi     = {10.1007/s44336-026-00036-1}
}

@article{ruan2023aadiff,
  title   = {{AADiff}: Audio-Aligned Video Synthesis with Text-to-Image Diffusion},
  author  = {Ruan, Yang and others},
  journal = {arXiv preprint arXiv:2305.04001},
  year    = {2023}
}

@inproceedings{zhang2023sadtalker,
  title     = {{SadTalker}: Learning Realistic 3D Motion Coefficients for Stylized Audio-Driven Single Image Talking Face Animation},
  author    = {Zhang, Wenxuan and Cun, Xiaodong and Wang, Xuan and Zhang, Yong and Shen, Xi and Guo, Yu and Shan, Ying and Wang, Fei},
  booktitle = {Proceedings of the IEEE/CVF Conference on Computer Vision and Pattern Recognition},
  year      = {2023}
}

@inproceedings{prajwal2020wav2lip,
  title     = {A Lip Sync Expert Is All You Need for Speech to Lip Generation In the Wild},
  author    = {Prajwal, K R and Mukhopadhyay, Rudrabha and Namboodiri, Vinay P and Jawahar, C V},
  booktitle = {Proceedings of the ACM International Conference on Multimedia},
  year      = {2020}
}

@inproceedings{shen2023difftalk,
  title     = {{DiffTalk}: Crafting Diffusion Models for Generalized Audio-Driven Portraits Animation},
  author    = {Shen, Shuai and Zhao, Wenliang and Meng, Zibin and Li, Wanhua and Zhu, Zheng and Zhou, Jie and Lu, Jiwen},
  booktitle = {Proceedings of the IEEE/CVF Conference on Computer Vision and Pattern Recognition},
  pages     = {1982--1991},
  year      = {2023}
}

@inproceedings{yu2023diffusionprior,
  title     = {Talking Head Generation with Probabilistic Audio-to-Visual Diffusion Priors},
  author    = {Yu, Zhentao and Yin, Zixin and Zhou, Deyu and Wang, Duomin and Wong, Finn and Wang, Baoyuan},
  booktitle = {Proceedings of the IEEE/CVF International Conference on Computer Vision},
  pages     = {7645--7655},
  year      = {2023}
}

@inproceedings{aneja2024facetalk,
  title     = {{FaceTalk}: Audio-Driven Motion Diffusion for Neural Parametric Head Models},
  author    = {Aneja, Shivangi and Thies, Justus and Dai, Angela and Nie{\ss}ner, Matthias},
  booktitle = {Proceedings of the IEEE/CVF Conference on Computer Vision and Pattern Recognition},
  pages     = {21263--21273},
  year      = {2024}
}

@inproceedings{peng2024synctalk,
  title     = {{SyncTalk}: The Devil Is in the Synchronization for Talking Head Synthesis},
  author    = {Peng, Ziqiao and Hu, Wentao and Shi, Yue and Zhu, Xiangyu and Zhang, Xiaomei and Zhao, Hao and He, Jun and Liu, Hongyan and Fan, Zhaoxin},
  booktitle = {Proceedings of the IEEE/CVF Conference on Computer Vision and Pattern Recognition},
  pages     = {666--676},
  year      = {2024}
}

@inproceedings{guan2025audcast,
  title     = {{AudCast}: Audio-Driven Human Video Generation by Cascaded Diffusion Transformers},
  author    = {Guan, Jiazhi and Wang, Kaisiyuan and Xu, Zhiliang and Yang, Quanwei and Sun, Yasheng and He, Shengyi and Liang, Borong and Cao, Yukang and Li, Yingying and Feng, Haocheng and Ding, Errui and Wang, Jingdong and Zhao, Youjian and Zhou, Hang and Liu, Ziwei},
  booktitle = {Proceedings of the IEEE/CVF Conference on Computer Vision and Pattern Recognition},
  pages     = {10678--10689},
  year      = {2025}
}

@inproceedings{corona2025vlogger,
  title     = {{VLOGGER}: Multimodal Diffusion for Embodied Avatar Synthesis},
  author    = {Corona, Enric and Zanfir, Andrei and Bazavan, Eduard Gabriel and Kolotouros, Nikos and Alldieck, Thiemo and Sminchisescu, Cristian},
  booktitle = {Proceedings of the IEEE/CVF Conference on Computer Vision and Pattern Recognition},
  pages     = {15896--15908},
  year      = {2025}
}

@inproceedings{cui2025hallo2,
  title     = {{Hallo2}: Long-Duration and High-Resolution Audio-Driven Portrait Image Animation},
  author    = {Cui, Jiahao and Li, Hui and Yao, Yao and Zhu, Hao and Shang, Hanlin and Cheng, Kaihui and Zhou, Hang and Zhu, Siyu and Wang, Jingdong},
  booktitle = {International Conference on Learning Representations},
  year      = {2025}
}

@inproceedings{lipman2023flowmatching,
  title     = {Flow Matching for Generative Modeling},
  author    = {Lipman, Yaron and Chen, Ricky T. Q. and Ben-Hamu, Heli and Nickel, Maximilian and Le, Matthew},
  booktitle = {International Conference on Learning Representations},
  year      = {2023}
}

@inproceedings{liu2023rectifiedflow,
  title     = {Flow Straight and Fast: Learning to Generate and Transfer Data with Rectified Flow},
  author    = {Liu, Xingchao and Gong, Chengyue and Liu, Qiang},
  booktitle = {International Conference on Learning Representations},
  year      = {2023}
}

@article{tjandra2025audiobox,
  title   = {Meta Audiobox Aesthetics: Unified Automatic Quality Assessment for Speech, Music, and Sound},
  author  = {Tjandra, Andros and Wu, Yi-Chiao and Guo, Baishan and Hoffman, John and Ellis, Brian and Vyas, Apoorv and Shi, Bowen and Chen, Sanyuan and Le, Matt and Zacharov, Nick and Wood, Carleigh and Lee, Ann and Hsu, Wei-Ning},
  journal = {arXiv preprint arXiv:2502.05139},
  year    = {2025}
}

@inproceedings{ke2021musiq,
  title     = {{MUSIQ}: Multi-Scale Image Quality Transformer},
  author    = {Ke, Junjie and Wang, Qifei and Wang, Yilin and Milanfar, Peyman and Yang, Feng},
  booktitle = {Proceedings of the IEEE/CVF International Conference on Computer Vision},
  pages     = {5148--5157},
  year      = {2021}
}

@inproceedings{chung2017syncnet,
  title     = {Out of Time: Automated Lip Sync in the Wild},
  author    = {Chung, Joon Son and Zisserman, Andrew},
  booktitle = {Asian Conference on Computer Vision Workshops},
  year      = {2017}
}

@article{yariv2023tempotokens,
  title   = {Diverse and Aligned Audio-to-Video Generation via Text-to-Video Model Adaptation},
  author  = {Yariv, Guy and Gat, Itai and Benaim, Sagie and Wolf, Lior and Schwartz, Idan and Adi, Yossi},
  journal = {arXiv preprint arXiv:2309.16429},
  year    = {2023}
}

@article{liu2026syncdpo,
  title   = {{SyncDPO}: Enhancing Temporal Synchronization in Video-Audio Joint Generation via Preference Learning},
  author  = {Liu, Ziwei and others},
  journal = {arXiv preprint arXiv:2605.12179},
  year    = {2026}
}

@article{chern2026speed,
  title={Speed by Simplicity: A Single-Stream Architecture for Fast Audio-Video Generative Foundation Model},
  author={Chern, Ethan and Teng, Hansi and Sun, Hanwen and Wang, Hao and Pan, Hong and Jia, Hongyu and Su, Jiadi and Li, Jin and Yu, Junjie and Liu, Lijie and others},
  journal={arXiv preprint arXiv:2603.21986},
  year={2026}
}

@article{fritsch1980monotone,
  title   = {Monotone Piecewise Cubic Interpolation},
  author  = {Fritsch, F. N. and Carlson, R. E.},
  journal = {SIAM Journal on Numerical Analysis},
  volume  = {17},
  number  = {2},
  pages   = {238--246},
  year    = {1980},
  doi     = {10.1137/0717021}
}

\clearpage
\appendix
\section*{Appendix}

This appendix is organized into three sections. We first present the complete TimeSteer sampling procedure and then derive Region-Aware Latent Remapping for speech intervals and non-speech gaps. Finally, we document the experimental setup and method implementations, explain how the localization components are selected, and provide expanded quantitative and qualitative results.

\section{TimeSteer Procedure}
\label{app:procedure}

Algorithm~\ref{alg:timesteer} summarizes how TimeSteer is integrated into standard flow-matching sampling. At each denoising step, the model first predicts the velocity $v_n$ and recovers the corresponding clean latent $\hat{x}_0^n$. TimeSteer then localizes the source spans and remaps $\hat{x}_0^n$ before converting it back to the velocity used by the original flow-matching update.

\begin{algorithm}[ht]
\caption{TimeSteer within flow-matching sampling}
\label{alg:timesteer}
\begin{algorithmic}[1]
\REQUIRE Frozen model $\mathcal{G}_\theta$, prompt $c$, and target intervals $\{\mathcal{T}^u\}_{u=1}^{N}$
\REQUIRE Initial noisy latent $x_M$ and noise schedule $\{\sigma_n\}_{n=0}^{M}$
\ENSURE Generated audio $A$ and video $V$
\FOR{$n=M,\ldots,1$}
    \STATE $v_n\gets\mathcal{G}_\theta(x_n,c,\sigma_n)$
    \STATE $\hat{x}_0^n\gets x_n-\sigma_n v_n$
    \STATE \textbf{Source Span Localization:} estimate $\{S^u\}_{u=1}^{N}$ from $\hat{x}_0^n$ and $c$
    \STATE \textbf{Region-Aware Latent Remapping:} construct $h$ from $\{(S^u,\mathcal{T}^u)\}_{u=1}^{N}$
    \FOR{each destination coordinate $t$}
        \STATE $k_t^-\gets\lfloor h(t)\rfloor,\quad k_t^+\gets\lceil h(t)\rceil$
        \STATE $\lambda_t\gets h(t)-k_t^-$
        \STATE $\tilde{x}_0^n(t)\gets
    (1-\lambda_t)\hat{x}_0^n(k_t^-)
        +\lambda_t\hat{x}_0^n(k_t^+)$
    \ENDFOR
    \STATE $\tilde{v}_n\gets
    (x_n-\tilde{x}_0^n)/\sigma_n$
    \STATE $x_{n-1}\gets
    x_n+(\sigma_{n-1}-\sigma_n)\tilde{v}_n$
\ENDFOR
\STATE $(A,V)\gets\operatorname{Decode}(x_0)$
\RETURN $(A,V)$
\end{algorithmic}
\end{algorithm}

The two TimeSteer operations in Algorithm~\ref{alg:timesteer} correspond to the successive stages described in the main paper. Specifically, Source Span Localization estimates $\{S^u\}_{u=1}^{N}$ from the text-to-audio cross-attention of $\hat{x}_0^n$, whereas Region-Aware Latent Remapping converts each target interval $\mathcal{T}^u$ to its destination speech interval and constructs the destination-to-source read map $h$. Once $h$ is obtained, the same neighboring indices and interpolation weights are applied to the audio and video branches of $\hat{x}_0^n$, preserving their temporal correspondence.

\section{Derivation of Region-Aware Latent Remapping}
\label{app:hybrid-theory}

Region-Aware Latent Remapping is designed around two distinct temporal requirements: preserving the internal speech structure within each utterance and smoothly connecting neighboring utterances across non-speech gaps. We first establish that the slope of the destination-to-source read map determines the local speaking-rate ratio. Based on this interpretation, we then derive the affine speech map that minimizes speaking-rate distortion and the cubic gap map that minimizes variation in the read-map slope.

\subsection{Speaking-Rate Interpretation of the Read-Map Slope}

Let $q_{\mathrm{src}}(t)$ and $q_{\mathrm{dst}}(t)$ denote the cumulative speech content rendered up to coordinate $t$ on the source and destination timelines, respectively. Their derivatives define the corresponding local speaking rates:
\begin{equation}
r_{\mathrm{src}}(t)
=
\frac{d q_{\mathrm{src}}(t)}{dt},
\qquad
r_{\mathrm{dst}}(t)
=
\frac{d q_{\mathrm{dst}}(t)}{dt}.
\label{eq:app-speaking-rates}
\end{equation}
The read map $h:[0,T-1]\rightarrow[0,T-1]$ specifies the source coordinate sampled at each destination coordinate, and therefore relates the two content functions as
\begin{equation}
q_{\mathrm{dst}}(t)
=
q_{\mathrm{src}}\!\left(h(t)\right).
\label{eq:app-content-remapping}
\end{equation}
Applying the chain rule gives
\begin{equation}
r_{\mathrm{dst}}(t)
=
r_{\mathrm{src}}\!\left(h(t)\right)h'(t).
\label{eq:app-rate-scaling}
\end{equation}
For a nonzero source speaking rate, this gives 
\begin{equation}
h'(t)
=
\frac{r_{\mathrm{dst}}(t)}
{r_{\mathrm{src}}\!\left(h(t)\right)}.
\label{eq:app-rate-ratio}
\end{equation}
Thus, $h'(t)$ gives the local multiplicative change from the original speaking rate to the output speaking rate. When $h'(t)=1$, the output speaking rate equals the original speaking rate. Since the same read map is applied to the audio and video latents, the associated visual articulation undergoes the same temporal scaling.

\subsection{Minimal-Distortion Remapping for Speech Intervals}

Within each speech interval, we aim to preserve the original speaking rate, corresponding to $h'(t)=1$. For utterance $u$, we therefore solve
\begin{equation}
\begin{aligned}
h_u^\star
&=
\arg\min_h\quad
\int_{d_0^u}^{d_1^u}
\bigl(h'(t)-1\bigr)^2\,dt,
\\
&\text{s.t.}\quad
h(d_0^u)=s_0^u,
\qquad
h(d_1^u)=s_1^u.
\end{aligned}
\label{eq:app-speech-variational}
\end{equation}
The Euler--Lagrange equation provides the stationary condition for this functional. Applying it directly to the integrand in Eq.~\eqref{eq:app-speech-variational} gives
\begin{equation}
\begin{aligned}
0
&=
\frac{\partial}{\partial h}
\bigl(h'(t)-1\bigr)^2
-
\frac{d}{dt}
\left[
\frac{\partial}{\partial h'}
\bigl(h'(t)-1\bigr)^2
\right]
\\
&=
-\frac{d}{dt}
\left[
2\bigl(h'(t)-1\bigr)
\right]
=
-2h''(t).
\end{aligned}
\label{eq:app-speech-el}
\end{equation}
The first partial derivative is zero because the integrand depends on $h'(t)$ but not directly on $h(t)$. Equation~\eqref{eq:app-speech-el} therefore requires
\begin{equation}
h''(t)=0,
\qquad
t\in[d_0^u,d_1^u].
\label{eq:app-speech-stationary}
\end{equation}
Integrating twice shows that every stationary solution has the affine form
\begin{equation}
h(t)=\alpha_u t+\beta_u.
\label{eq:app-speech-affine-general}
\end{equation}

We determine the two constants from the endpoint constraints:
\begin{equation}
\alpha_u d_0^u+\beta_u=s_0^u,
\qquad
\alpha_u d_1^u+\beta_u=s_1^u.
\label{eq:app-speech-linear-system}
\end{equation}
Subtracting the first equation from the second gives
\begin{equation}
\alpha_u
=
\frac{s_1^u-s_0^u}{d_1^u-d_0^u}
=
\kappa_u,
\qquad
\beta_u=s_0^u-\kappa_u d_0^u.
\label{eq:app-speech-parameters}
\end{equation}
Substituting these constants into Eq.~\eqref{eq:app-speech-affine-general} yields
\begin{equation}
h_u^\star(t)
=
s_0^u
+
\kappa_u
\bigl(t-d_0^u\bigr),
\qquad
t\in[d_0^u,d_1^u].
\label{eq:app-minimal-distortion-map}
\end{equation}

It remains to determine the type of this stationary solution. The second derivative of the integrand with respect to $h'$ is
\begin{equation}
\frac{\partial^2}{\partial(h')^2}
\bigl(h'(t)-1\bigr)^2
=
2
>
0.
\label{eq:app-speech-convexity}
\end{equation}
Thus, the integrand is strictly convex in $h'$. Together with the affine endpoint constraints, this makes the complete objective strictly convex over the feasible read maps. Consequently, the stationary affine map in Eq.~\eqref{eq:app-minimal-distortion-map} is the unique global minimizer of Eq.~\eqref{eq:app-speech-variational}.

\subsection{Curvature Bridging for Non-speech Gaps}

Across a non-speech gap, the speaking-rate ratios inherited from neighboring speech intervals may differ. As established above, $h'(t)$ is the local speaking-rate ratio, i.e., the output speaking rate divided by the original speaking rate, so $h''(t)=dh'(t)/dt$ measures how rapidly this ratio changes over destination time. We use the gap to transition smoothly between the two ratios by minimizing the accumulated squared magnitude of $h''(t)$ under the boundary constraints.

Consider a nonempty interior gap between utterances $u$ and $u+1$, where $d_1^u<d_0^{u+1}$. Its boundary slopes are inherited directly from the adjacent affine speech maps as $\kappa_u$ and $\kappa_{u+1}$. Following the formulation in the main paper, we solve
\begin{equation}
\begin{aligned}
h_{u,u+1}^\star
&=
\arg\min_h\quad
\int_{d_1^u}^{d_0^{u+1}}
\bigl(h''(t)\bigr)^2\,dt,
\\
&\text{s.t.}\quad
h(d_1^u)=s_1^u,
\quad
h(d_0^{u+1})=s_0^{u+1},
\\
&\phantom{\text{s.t.}\quad}
h'(d_1^u)=\kappa_u,
\quad
h'(d_0^{u+1})=\kappa_{u+1}.
\end{aligned}
\label{eq:app-gap-variational}
\end{equation}

To derive the stationary condition, we perturb a candidate read map $h$ in an arbitrary direction $\eta$:
\begin{equation}
h_\epsilon(t)
=
h(t)+\epsilon\eta(t),
\label{eq:app-gap-perturbation}
\end{equation}
where $\epsilon$ is a scalar controlling the perturbation magnitude and $\eta(t)$ specifies its shape over the gap. The perturbed map must satisfy the same four boundary constraints as $h$. Substituting Eq.~\eqref{eq:app-gap-perturbation} into these constraints shows that $\eta$ must satisfy
\begin{equation}
\begin{aligned}
\eta(d_1^u)&=0,
\qquad
\eta(d_0^{u+1})=0,
\\
\eta'(d_1^u)&=0,
\qquad
\eta'(d_0^{u+1})=0.
\end{aligned}
\label{eq:app-gap-perturbation-boundary}
\end{equation}
The first line keeps the two endpoint values of the read map fixed, while the second keeps its two boundary slopes fixed. Any sufficiently smooth $\eta$ satisfying these conditions is an admissible perturbation.

Let
\begin{equation}
J[h]
=
\int_{d_1^u}^{d_0^{u+1}}
\bigl(h''(t)\bigr)^2\,dt
\label{eq:app-gap-objective}
\end{equation}
denote the bending-energy objective. Under the perturbation in Eq.~\eqref{eq:app-gap-perturbation}, it becomes
\begin{equation}
J[h_\epsilon]
=
\int_{d_1^u}^{d_0^{u+1}}
\bigl(h''(t)+\epsilon\eta''(t)\bigr)^2\,dt.
\label{eq:app-gap-perturbed-objective}
\end{equation}
If $h$ is a minimizer, then $\epsilon=0$ must be a stationary point of $J[h_\epsilon]$ for every admissible direction $\eta$. Therefore,
\begin{equation}
\left.
\frac{dJ[h_\epsilon]}{d\epsilon}
\right|_{\epsilon=0}
=
2\int_{d_1^u}^{d_0^{u+1}}
h''(t)\eta''(t)\,dt
=
0.
\label{eq:app-gap-first-variation}
\end{equation}

We now integrate the remaining integral by parts twice. Each integration transfers one derivative from $\eta$ to $h$, and all boundary terms vanish because both $\eta$ and $\eta'$ are zero at the two endpoints. Therefore,
\begin{equation}
\int_{d_1^u}^{d_0^{u+1}}
h''(t)\eta''(t)\,dt
=
\int_{d_1^u}^{d_0^{u+1}}
h^{(4)}(t)\eta(t)\,dt.
\label{eq:app-gap-integration-by-parts}
\end{equation}
Combining Eqs.~\eqref{eq:app-gap-first-variation} and \eqref{eq:app-gap-integration-by-parts} gives
\begin{equation}
\int_{d_1^u}^{d_0^{u+1}}
h^{(4)}(t)\eta(t)\,dt
=
0
\end{equation}
for every admissible perturbation $\eta$. This class includes arbitrary smooth perturbations supported within the interior of the gap. The fundamental lemma of the calculus of variations therefore gives the Euler--Lagrange condition
\begin{equation}
h^{(4)}(t)=0.
\label{eq:app-gap-el}
\end{equation}
Therefore, every stationary solution is a polynomial of degree at most three.

To obtain its coefficients, define
\begin{equation}
\begin{aligned}
L_u
&=
d_0^{u+1}-d_1^u,
\\
\Delta_u
&=
s_0^{u+1}-s_1^u,
\\
\xi
&=
t-d_1^u.
\end{aligned}
\label{eq:app-gap-notation}
\end{equation}
Writing
\begin{equation}
h_{u,u+1}^\star(t)
=
\sum_{r=0}^{3}a_{u,r}\xi^r
\label{eq:app-gap-solution}
\end{equation}
and applying the four constraints in Eq.~\eqref{eq:app-gap-variational} gives
\begin{equation}
\begin{aligned}
a_{u,0}
&=
s_1^u,
\\
a_{u,1}
&=
\kappa_u,
\\
a_{u,2}
&=
\frac{3\Delta_u}{L_u^2}
-
\frac{2\kappa_u+\kappa_{u+1}}{L_u},
\\
a_{u,3}
&=
-\frac{2\Delta_u}{L_u^3}
+
\frac{\kappa_u+\kappa_{u+1}}{L_u^2}.
\end{aligned}
\label{eq:app-gap-coefficients}
\end{equation}
These coefficients ensure that both $h$ and $h'$ match the adjacent affine speech maps at the two gap boundaries. The resulting cubic-polynomial bridge therefore makes the read map continuously differentiable across speech intervals and non-speech gaps.

Finally, the second variation is
$2\int_{d_1^u}^{d_0^{u+1}}(\eta''(t))^2dt$,
which is strictly positive for every nonzero admissible perturbation. Indeed, $\eta''=0$ would make $\eta$ affine, and the homogeneous endpoint constraints would then force $\eta\equiv0$. The cubic-polynomial solution is therefore the unique global minimizer.

\section{Additional Experimental Details and Results}
\label{app:additional-experiments}

This section provides the information required to reproduce SpeechShift evaluation, specifies how each training-free baseline is implemented on the joint audio-visual backbones, and reports additional results.

\subsection{Reproducibility Details}
\label{app:reproducibility-details}

\paragraph{Data specification and release.}
Each SpeechShift entry contains a scene prompt, an ordered list of quoted utterances, the speaker associated with each utterance, its begin--end target interval, and metadata specifying the scene, speaking structure, acoustic condition, and visual-action condition. The benchmark contains 400 prompts, 600 utterance-level targets, and 102 scenes, with 100 prompts in each of the four speaker--utterance structures described in the main paper. Upon publication, we will release SpeechShift and its construction metadata under a license permitting free use for research.

\paragraph{Code release.}
Upon publication, we will release the source code for TimeSteer, benchmark construction, baseline adaptation, and evaluation under a license permitting free use for research. The released implementation will document the correspondence between its principal stages and the equations and pseudocode in the paper.

\paragraph{Experimental setup.}
All experiments are conducted in Python~3.10 on NVIDIA RTX A6000 GPUs. Unless otherwise stated, the main comparison uses one run per method with a fixed random seed of 42; the cross-seed robustness study uses five seeds as specified below. All methods generate 5-second audio-visual clips: LTX-2 uses 24 FPS, whereas daVinci-MagiHuman uses 25 FPS. To accommodate the available GPU memory and compute budget, we use the FP8-quantized distilled configuration of each backbone with its eight-step sampler. Within each backbone, all methods use the same released checkpoint, prompts, target intervals, native inference resolution, numerical precision, and guidance scale. No model fine-tuning or test-time optimization is performed.
The configurations introduced by each competing control method are listed below; all remaining backbone parameters retain their official defaults.

\subsection{Method Implementations}
\label{app:baseline-implementations}

\subsubsection{Uncontrolled}

The Uncontrolled baseline applies the backbone's standard sampling procedure to the original scene prompt. It receives no target-interval information and introduces no intervention during denoising.

\subsubsection{Textual Timing}

Textual Timing communicates the target schedule only through the conditioning prompt. For each quoted utterance, its begin and end timestamps are converted into an explicit natural-language timing instruction and appended to the original scene description. For example, the original clause \emph{a mechanic says: ``please sign here on the line''} becomes \emph{a mechanic says: ``please sign here on the line'' from 1.6\,s to 3.4\,s}. Sampling otherwise remains unchanged, with no modification to attention or latent representations.

\subsubsection{FreeAudio}
\label{app:freeaudio}

\paragraph{Original method.}
FreeAudio is a training-free framework for timing-controlled long-form text-to-audio generation. It divides a timing-annotated prompt into non-overlapping windows and rewrites the desired audio event in each window as a window-specific description. Its Decoupling and Aggregating Attention Control (DAAC) then binds each description to its window by suppressing cross-attention between the corresponding text tokens and audio queries outside that window. The window-controlled attention responses are subsequently aggregated with the original responses to balance temporal control and generation quality.

\paragraph{SpeechShift adaptation.}
SpeechShift represents each scheduling request as a quoted utterance paired with a target interval. For each utterance, we identify the first and last token indices of the quoted utterance and map the begin and end timestamps of its target interval to the corresponding audio and video latent coordinates. For LTX-2, DAAC is applied to both text-to-audio and text-to-video cross-attention. For daVinci-MagiHuman, the same constraint is applied within unified attention to the interactions between quoted-text keys and audio or video queries. In both architectures, the audio and video coordinates are derived from the same target timestamps, providing a common temporal constraint for the two modalities.

\paragraph{Implementation.}
For LTX-2, we apply DAAC at all eight denoising steps and in transformer blocks $20$--$30$, using all attention heads. The DAAC mask is linearly ramped over four query-token positions at each target-window boundary. Let $A_{\mathrm{base}}$ denote the original attention probabilities and $A_{\mathrm{DAAC}}$ the probabilities after applying the window constraint. Following FreeAudio's aggregation principle, we compute
\begin{equation}
A_{\mathrm{out}}
=
\alpha A_{\mathrm{base}}
+
(1-\alpha)A_{\mathrm{DAAC}},
\label{eq:freeaudio-aggregation}
\end{equation}
where $\alpha$ controls the contribution of the original attention. We set $\alpha=0.2$ and apply $A_{\mathrm{out}}$ to both audio and video queries.

For daVinci-MagiHuman, we apply DAAC at every denoising step and in transformer blocks $10$--$39$, using all attention heads. We use a two-query-token boundary ramp and set $\alpha=0$, so $A_{\mathrm{out}}=A_{\mathrm{DAAC}}$. The additive DAAC bias applied to the attention logits is scaled by $1.5$, controlling the strength of the window constraint. We also set the query/key (Q/K) steering strength to $1.5$, which amplifies audio and video queries inside the target interval together with the keys of the corresponding quoted tokens.

\subsubsection{Prompt Relay}
\label{app:promptrelay}

\paragraph{Original method.}
Prompt Relay is a training-free method for temporally controlling multiple events in text-to-video generation. It assigns each local prompt to a temporal segment while retaining a global prompt throughout the video. Its Boundary-Attention Decay subtracts a distance-dependent penalty from cross-attention logits, allowing queries inside a segment to attend primarily to the corresponding local prompt while varying the penalty smoothly near segment boundaries.
Specifically, the controlled attention is computed as
\begin{equation}
\operatorname{Attn}(Q,K,V)
=
\operatorname{softmax}\!\left(
\frac{QK^\top}{\sqrt{d}}-C
\right)V,
\label{eq:pr-attn}
\end{equation}
where $Q$, $K$, and $V$ are the query, key, and value matrices, $d$ is the key dimension, and $C$ is the Boundary-Attention Decay penalty. For a local prompt assigned to segment $s$, the penalty is
\begin{equation}
C_{ij}
=
\begin{cases}
\dfrac{\bigl(|i-m_s|-w_s\bigr)_+^2}
{2\sigma_s^2},
& j\in\mathcal{K}_s,\\[6pt]
0,
& j\notin\mathcal{K}_s,
\end{cases}
\label{eq:pr-penalty}
\end{equation}
where $i$ and $j$ index the temporal query and text key, $\mathcal{K}_s$ contains the text keys of the local prompt, $m_s$ is the midpoint of its assigned segment, $w_s$ specifies the central region with no penalty, and $(z)_+=\max(z,0)$. The scale $\sigma_s$ controls how rapidly the penalty increases beyond this region. Thus, attention to the local prompt remains unrestricted near its assigned segment and is progressively suppressed as the temporal distance increases.

\paragraph{SpeechShift adaptation.}
SpeechShift contains a single scene prompt with one or more quoted utterances. We treat the token range of each quoted utterance as a local prompt and its target interval as the corresponding temporal segment, while the remaining scene description provides global context. The begin and end timestamps are mapped to both audio and video latent coordinates. For LTX-2, the resulting penalty is applied to text-to-audio and text-to-video cross-attention. For daVinci-MagiHuman, it is applied within unified attention to the interactions between quoted-text keys and audio or video queries. This assigns the same utterance-level schedule to both modalities.

\paragraph{Implementation.}
For LTX-2, we apply Boundary-Attention Decay at all eight denoising steps and in transformer blocks $10$--$40$, using all attention heads. A separate penalty is constructed for each quoted token range and its target audio and video coordinates. The penalties vary smoothly with distance from the target interval. We set $\varepsilon=0.1$, where $\varepsilon$ specifies the attention factor retained at the segment boundary and a smaller value produces stronger decay.

For daVinci-MagiHuman, we apply the penalty at every denoising step and in transformer blocks $10$--$39$, using all attention heads. We set $\varepsilon=0.05$ and multiply the attention-logit penalty by $1.5$. The pre-attention steering strength is also set to $1.5$, controlling how strongly audio and video queries are downweighted according to the same temporal penalty.

\subsubsection{TimeSteer}
\label{app:timesteer-implementation}

\paragraph{Original method.}
Our TimeSteer is a training-free framework for inference-time speech scheduling in joint audio-visual diffusion models. It first performs Source Span Localization, which recovers each quoted utterance's model-implied source interval from a timing-sensitive attention response of the predicted clean latent. Region-Aware Latent Remapping then constructs a destination-to-source read map that relocates the corresponding audio-visual latent content to the requested target interval. Affine speech segments preserve the internal temporal structure of each utterance, while cubic bridges connect neighboring segments smoothly across non-speech gaps. The remapped clean latent is returned to the original sampler, allowing subsequent denoising steps to refine the relocated content without retraining the backbone.

\paragraph{SpeechShift adaptation.}
Each SpeechShift input consists of a scene prompt and one or more quoted utterances paired with begin--end target intervals. We identify the conditioning-token range of every quoted utterance and convert its target timestamps to the corresponding audio and video latent coordinates. At each denoising step, TimeSteer independently localizes a source span for every utterance and pairs it with its target interval to construct a single temporally ordered read map. For LTX-2, localization uses text-to-audio cross-attention from its separate audio stream, and the common wall-clock read map is evaluated on both the audio and video latent grids. For daVinci-MagiHuman, localization uses the interactions between quoted-text keys and audio queries in unified attention, and the same map is applied to the audio and video token groups. In both backbones, the original decoder produces the final 5-second audio-visual clip, with no post-hoc waveform or video editing.

\paragraph{Implementation.}
TimeSteer is applied at all eight denoising steps. At step $n$, we recover the predicted clean latent as $\hat{x}_0^n=x_n-\sigma_n v_n$ and replay it through the frozen transformer solely to expose the selected attention responses; this replay requires neither gradients nor parameter updates. For each utterance, responses over its quoted tokens are summed at every audio latent position, and the source span is bounded by the first and last positions whose response reaches the relative threshold $\tau=0.5$. LTX-2 uses transformer layer 25 and attention head 30. For daVinci-MagiHuman, we aggregate eight layer--head pairs: $(29,24)$, $(29,20)$, $(31,26)$, $(31,28)$, $(31,29)$, $(34,22)$, $(34,20)$, and $(35,29)$.

The localized source--target pairs define affine maps within speech intervals and cubic bridges across non-speech gaps, with the clip endpoints fixed. We evaluate the resulting continuous map on each discrete modality grid and linearly interpolate neighboring source positions. Finally, the scheduled clean latent $\tilde{x}_0^n$ is converted back to $\tilde{v}_n=(x_n-\tilde{x}_0^n)/\sigma_n$, and the backbone proceeds with its original sampler update. Region-Aware Latent Remapping introduces no tuned parameter beyond the localized source spans and user-specified target intervals.

\subsection{Localization Component Selection}
\label{app:localization-component-selection}

\begin{figure*}[ht]
\centering
\includegraphics[width=0.22\linewidth]{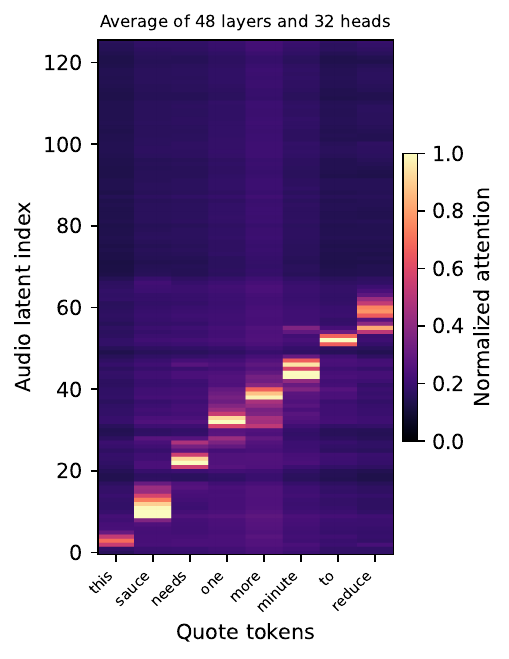}\hspace{0.035\linewidth}
\includegraphics[width=0.70\linewidth]{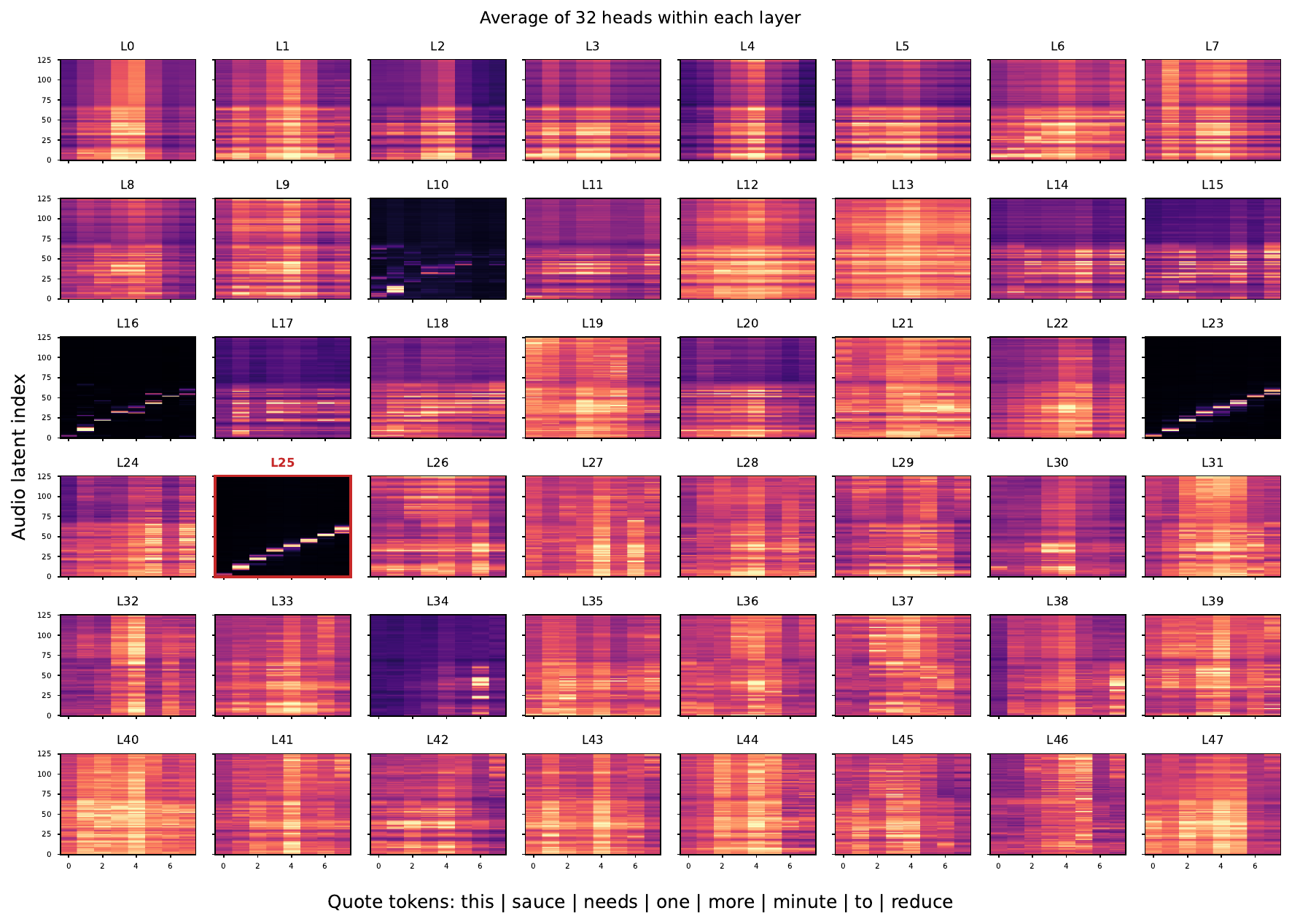}
\caption{Layer selection using LTX-2 text-to-audio attention on the predicted clean latent at denoise step index 5. The left panel averages text-to-audio attention over all layers and heads; the right panel averages the heads within each layer. Layer 25 is retained for head-level inspection.}
\label{fig:head-selection-ltx-overview}
\end{figure*}

\begin{figure*}[ht]
\centering
\includegraphics[width=0.22\linewidth]{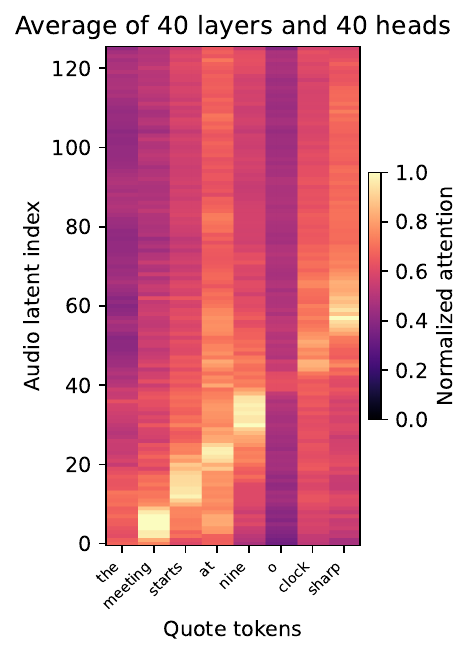}\hspace{0.035\linewidth}
\includegraphics[width=0.70\linewidth]{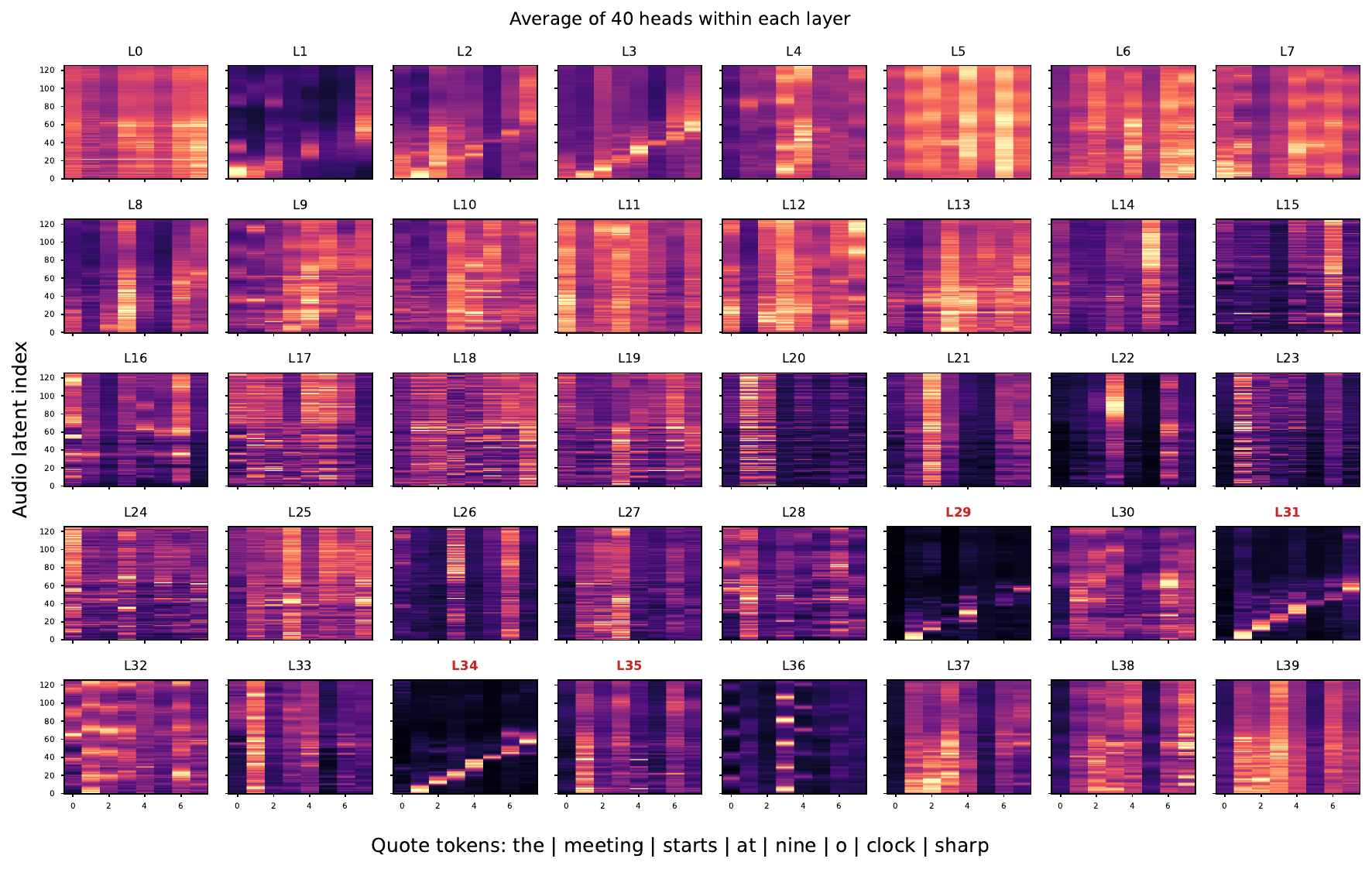}
\caption{Layer selection using daVinci-MagiHuman text-to-audio attention on the predicted clean latent at denoise step index 5. The left panel averages text-to-audio attention over all layers and heads; the right panel averages the heads within each layer. Layers 29, 31, 34, and 35 are retained for head-level inspection.}
\label{fig:head-selection-davinci-overview}
\end{figure*}

\begin{figure*}[ht]
\centering
\includegraphics[width=\linewidth]{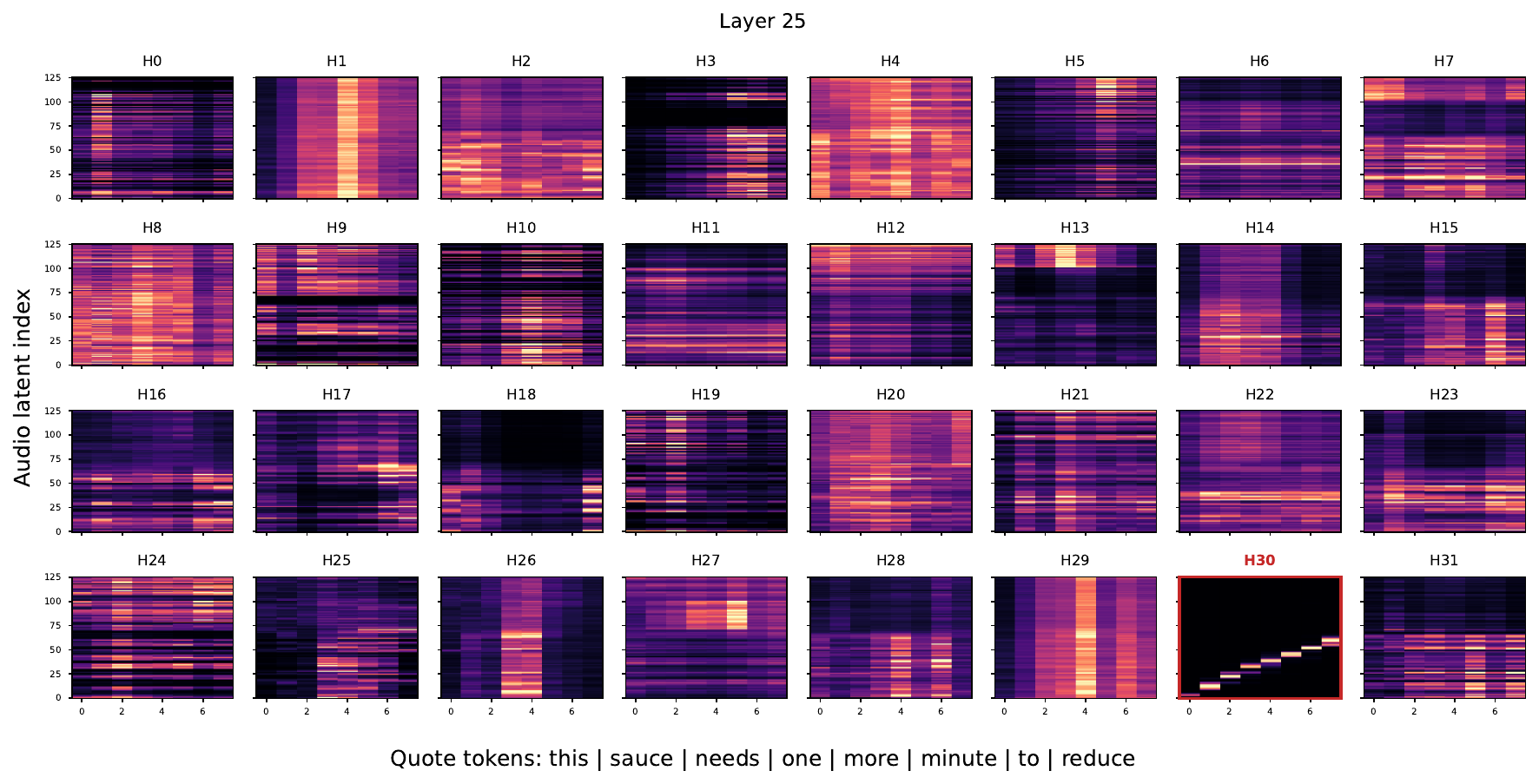}
\caption{Head-level text-to-audio attention on the predicted clean latent in LTX-2 layer 25. The selected head 30 is outlined and labeled in red.}
\label{fig:head-selection-ltx-layer25}
\end{figure*}

\begin{figure*}[ht]
\centering
\includegraphics[width=\linewidth]{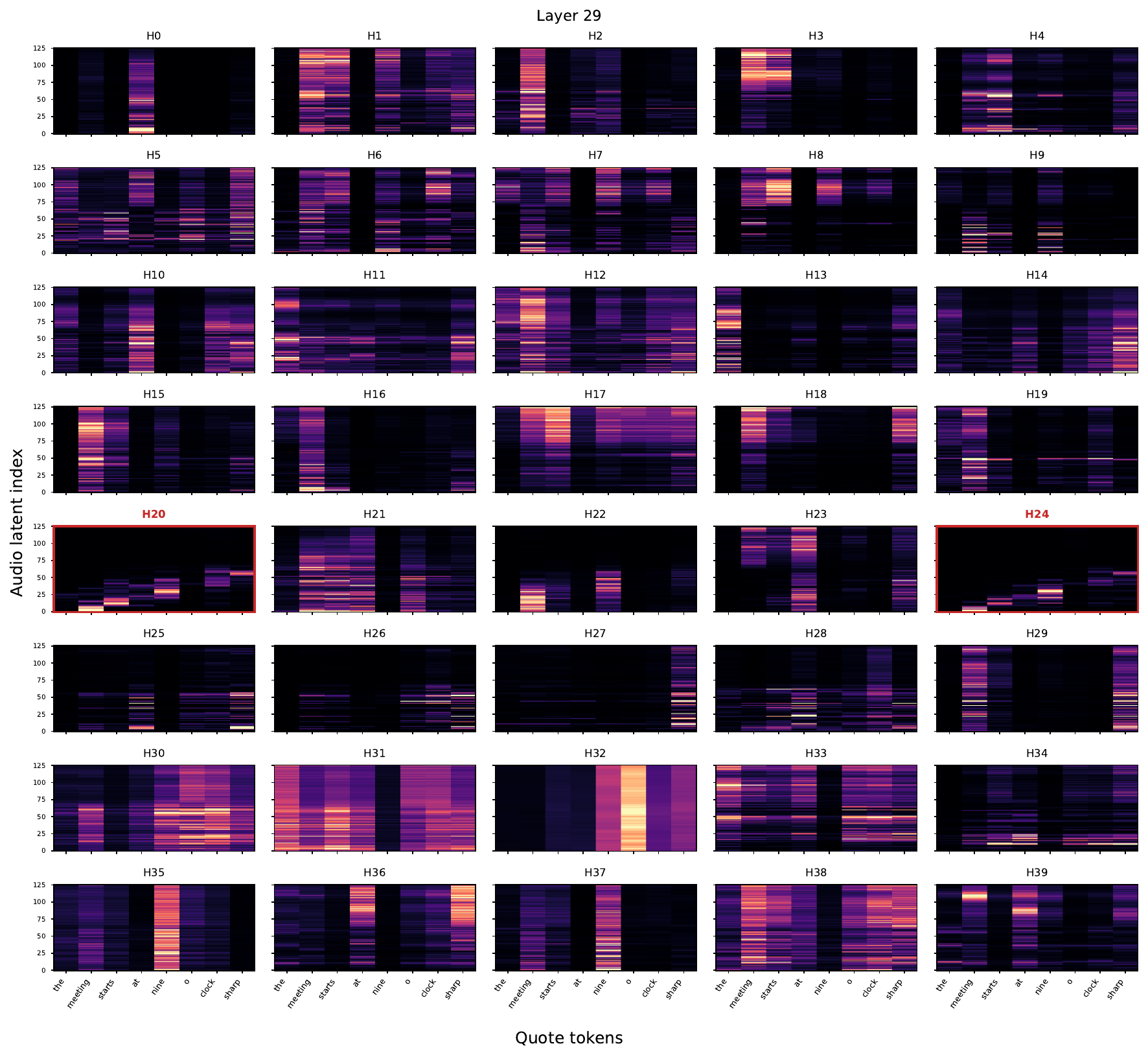}
\caption{Head-level text-to-audio attention on the predicted clean latent in daVinci-MagiHuman layer 29. Selected heads 20 and 24 are outlined and labeled in red.}
\label{fig:head-selection-davinci-layer29}
\end{figure*}

\begin{figure*}[ht]
\centering
\includegraphics[width=\linewidth]{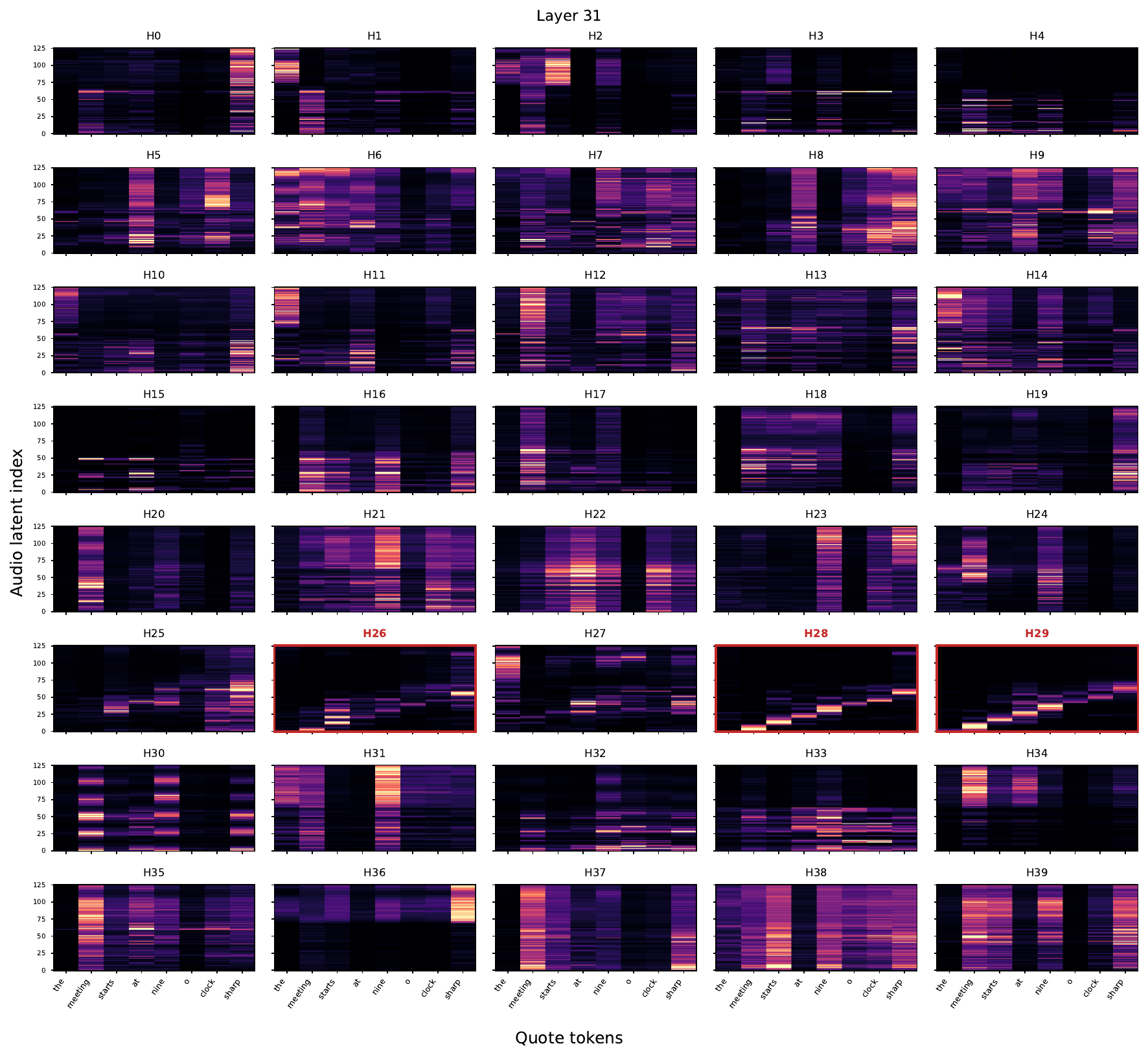}
\caption{Head-level text-to-audio attention on the predicted clean latent in daVinci-MagiHuman layer 31. Selected heads 26, 28, and 29 are outlined and labeled in red.}
\label{fig:head-selection-davinci-layer31}
\end{figure*}

\begin{figure*}[ht]
\centering
\includegraphics[width=\linewidth]{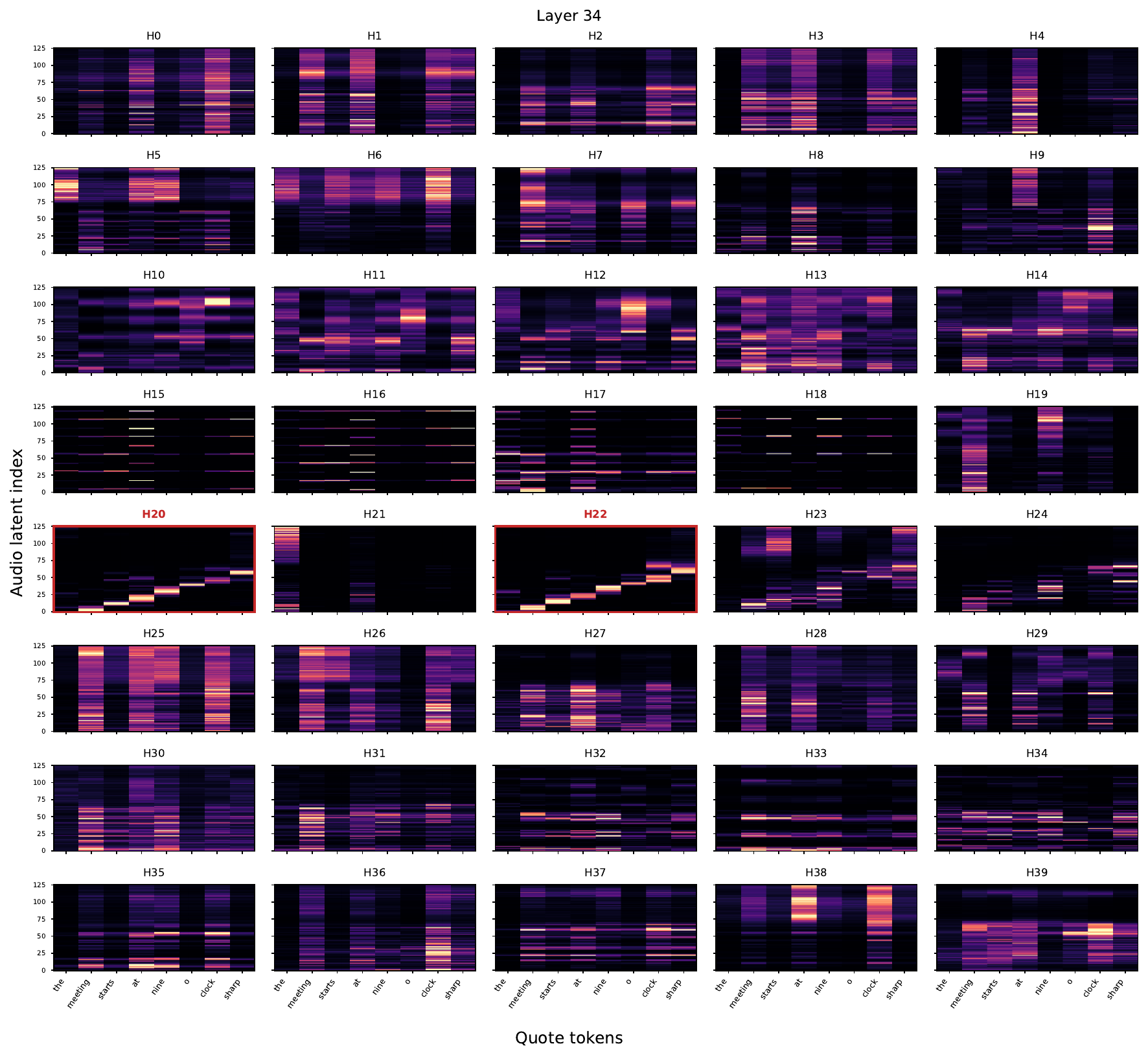}
\caption{Head-level text-to-audio attention on the predicted clean latent in daVinci-MagiHuman layer 34. Selected heads 20 and 22 are outlined and labeled in red.}
\label{fig:head-selection-davinci-layer34}
\end{figure*}

\begin{figure*}[ht]
\centering
\includegraphics[width=\linewidth]{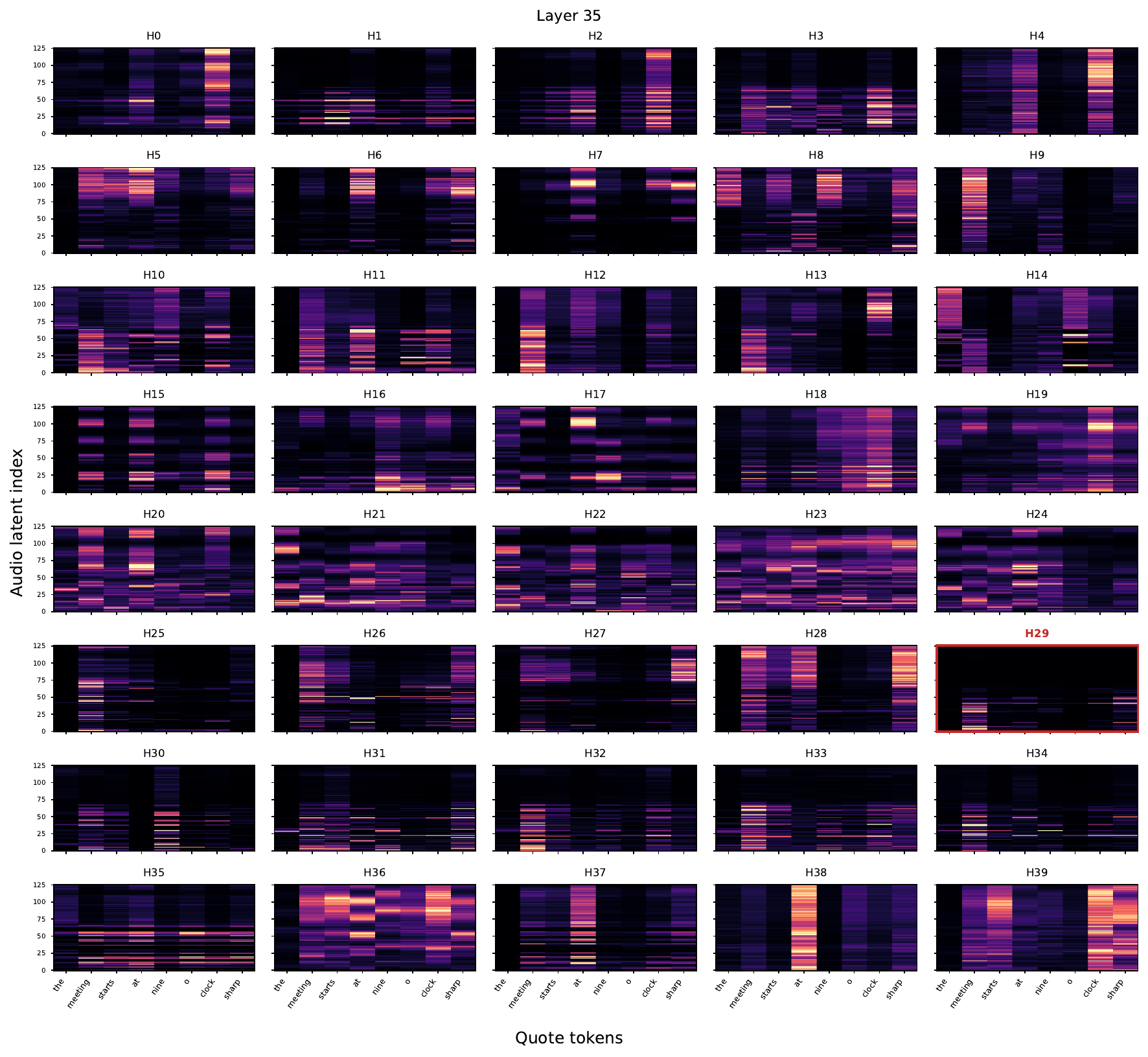}
\caption{Head-level text-to-audio attention on the predicted clean latent in daVinci-MagiHuman layer 35. Selected head 29 is outlined and labeled in red.}
\label{fig:head-selection-davinci-layer35}
\end{figure*}

We identify localization components through a hierarchical analysis at progressively finer levels of attention aggregation. Averaging text-to-audio attention across the full transformer reveals a temporally diffuse but discernible concentration within a bounded audio interval. Layer-wise aggregation shows that this temporal structure is substantially sharper in a small subset of layers, as illustrated for LTX-2 and daVinci-MagiHuman in Figures~\ref{fig:head-selection-ltx-overview} and~\ref{fig:head-selection-davinci-overview}, respectively. Resolving these layers into individual heads further isolates the most temporally concentrated responses, as shown in Figure~\ref{fig:head-selection-ltx-layer25} for LTX-2 and Figures~\ref{fig:head-selection-davinci-layer29}--\ref{fig:head-selection-davinci-layer35} for daVinci-MagiHuman.

The analysis isolates layer 25 and head 30 in LTX-2, and eight heads across layers 29, 31, 34, and 35 in daVinci-MagiHuman. Its applicability to both separate cross-attention and unified-attention backbones supports the use of a common selection procedure across architectures rather than model-specific heuristics.

\subsection{Additional Results}
\label{app:complete-quality}

\subsubsection{Complete Metric Comparison}
\label{app:complete-metric-comparison}

Table~\ref{tab:quality-full} extends the representative comparison in the main paper with five additional metrics. CLAP measures audio--text alignment, CLVP measures quote-level video--text consistency, while MANIQA and LAION provide complementary no-reference assessments of visual quality. LSE-C and LSE-D measure audio--visual synchronization in terms of confidence and feature distance, respectively. Both metrics are averaged over target intervals with valid face detections. We report latency as the total generation time divided by the eight denoising steps.

\begin{table*}[ht]
\centering
\footnotesize
\setlength{\tabcolsep}{0.5pt}
\begin{tabular*}{\textwidth}{@{\extracolsep{\fill}}lcccc|ccc|cccc|cc|c@{}}
\toprule
&
\multicolumn{4}{c|}{\textbf{Interval Controllability}}
&
\multicolumn{3}{c|}{\textbf{Semantic Fidelity}}
&
\multicolumn{4}{c|}{\textbf{Perceptual Quality}}
&
\multicolumn{2}{c|}{\textbf{Cross-modal}}
&
\multicolumn{1}{c}{\textbf{Efficiency}}\\
\cmidrule(lr){2-5}
\cmidrule(lr){6-8}
\cmidrule(lr){9-12}
\cmidrule(lr){13-14}
\cmidrule(l){15-15}
\textbf{Method}
& \textbf{HR$_{0.2}$}$\uparrow$
& \textbf{IoU}$\uparrow$
& \textbf{$mE_s$ (s)}$\downarrow$
& \textbf{$mE_e$ (s)}$\downarrow$
& \textbf{WER}$\downarrow$
& \textbf{CLAP}$\uparrow$
& \textbf{CLVP}$\uparrow$
& \textbf{PQ}$\uparrow$
& \textbf{MUSIQ}$\uparrow$
& \textbf{MANIQA}$\uparrow$
& \textbf{LAION}$\uparrow$
& \textbf{LSE-C}$\uparrow$
& \textbf{LSE-D}$\downarrow$
& \textbf{Latency (s)}$\downarrow$\\
\midrule
\multicolumn{15}{l}{\textbf{LTX-2}}\\
Uncontrolled & 0.21 & 0.63 & 0.56 & 0.54 & \textbf{0.05} & 0.21 & 0.35 & 5.36 & 52.26 & 0.27 & 4.44 & 3.13 & 11.60 & 3.47 \\
Textual Timing & 0.19 & 0.60 & 0.59 & 0.59 & 0.07 & \textbf{0.22} & 0.35 & 5.36 & 52.26 & 0.27 & \textbf{4.47} & \textbf{3.18} & 11.55 & \textbf{3.44} \\
FreeAudio & 0.19 & 0.54 & 0.49 & 0.67 & 0.35 & 0.20 & 0.35 & \textbf{5.92} & \textbf{52.55} & \textbf{0.27} & 4.44 & 2.92 & 11.08 & 3.49 \\
Prompt Relay & 0.31 & 0.68 & 0.27 & 0.35 & 0.20 & 0.21 & 0.35 & 5.52 & 52.00 & 0.27 & 4.45 & 3.06 & 11.13 & 3.45 \\
\rowcolor{gray!10}
TimeSteer & \textbf{0.73} & \textbf{0.87} & \textbf{0.11} & \textbf{0.15} & 0.07 & 0.22 & \textbf{0.35} & 5.32 & 52.51 & 0.27 & 4.45 & 3.10 & \textbf{10.95} & 3.84 \\
\addlinespace
\multicolumn{15}{l}{\textbf{daVinci-MagiHuman}}\\
Uncontrolled & 0.09 & 0.63 & 0.63 & 0.49 & \textbf{0.06} & 0.23 & 0.35 & 5.53 & 69.36 & 0.38 & 4.92 & 3.67 & 9.82 & 6.60 \\
Textual Timing & 0.08 & 0.61 & 0.64 & 0.53 & 0.09 & 0.23 & 0.35 & 5.50 & 68.98 & 0.38 & 4.93 & \textbf{3.87} & 9.57 & \textbf{6.45} \\
FreeAudio & 0.01 & 0.16 & 1.04 & 1.40 & 0.95 & 0.20 & \textbf{0.36} & 5.24 & 66.80 & 0.34 & 4.83 & 1.63 & \textbf{9.47} & 8.19 \\
Prompt Relay & 0.21 & 0.68 & 0.39 & 0.32 & 0.29 & \textbf{0.24} & 0.35 & \textbf{5.61} & \textbf{70.87} & \textbf{0.44} & \textbf{4.95} & 3.66 & 10.07 & 10.52 \\
\rowcolor{gray!10}
TimeSteer & \textbf{0.53} & \textbf{0.79} & \textbf{0.19} & \textbf{0.18} & 0.08 & 0.23 & 0.35 & 5.42 & 69.08 & 0.38 & 4.91 & 3.56 & 9.70 & 6.57 \\
\bottomrule
\end{tabular*}
\caption{Complete SpeechShift results. Best results per backbone are in bold. $\uparrow$/$\downarrow$ indicates higher/lower is better.}
\label{tab:quality-full}
\end{table*}

Across both backbones, TimeSteer more than doubles the best baseline hit rate ($0.73$ versus $0.31$ on LTX-2 and $0.53$ versus $0.21$ on daVinci-MagiHuman) while also improving temporal IoU and boundary accuracy. These gains preserve generated content: WER remains low, and semantic-fidelity, perceptual-quality, and cross-modal scores remain comparable to uncontrolled generation and the strongest baselines.

The additional cost is small. TimeSteer's amortized per-step latency remains close to uncontrolled generation ($3.84$ versus $3.47$\,s on LTX-2 and $6.57$ versus $6.60$\,s on daVinci-MagiHuman). Unlike attention-based baselines that inject masks or logit penalties into daVinci-MagiHuman's unified attention, TimeSteer remaps the predicted clean latent and remains compatible with accelerated attention operators.

\subsubsection{Cross-Seed Robustness across Speaking Structures}
\label{app:speaking-structure-results}

To evaluate seed robustness across speaking structures, we select 20 SpeechShift cases and run each method with five different seeds. Table~\ref{tab:structure-cross-seed} reports mean$\pm$std for the four structure types, with LTX-2 and daVinci-MagiHuman shown side by side.

Across all four speaking structures, TimeSteer achieves the highest hit rate on both backbones. It also gives the highest IoU in every LTX-2 setting and the highest or tied-highest IoU in every daVinci-MagiHuman setting. WER remains competitive: TimeSteer matches the best result in two LTX-2 settings and one daVinci-MagiHuman setting, and stays within $0.01$--$0.04$ of the best result elsewhere. These gains persist across five seeds, showing that the improvement in temporal control is not tied to a particular random seed.

\begin{table*}[ht]
\centering
\begin{tabular*}{\textwidth}{@{\extracolsep{\fill}}lcccccc@{}}
\toprule
& \multicolumn{3}{c}{\textbf{LTX-2}}
& \multicolumn{3}{c}{\textbf{daVinci-MagiHuman}}\\
\cmidrule(lr){2-4}\cmidrule(lr){5-7}
\textbf{Method}
& \textbf{HR$_{0.2}$}$\uparrow$ & \textbf{IoU}$\uparrow$ & \textbf{WER}$\downarrow$
& \textbf{HR$_{0.2}$}$\uparrow$ & \textbf{IoU}$\uparrow$ & \textbf{WER}$\downarrow$\\
\midrule
\multicolumn{7}{l}{\textit{Single speaker, single utterance}}\\
Uncontrolled
& 0.12$\pm$0.11 & 0.52$\pm$0.01 & \textbf{0.03$\pm$0.00}
& 0.00$\pm$0.00 & 0.46$\pm$0.04 & \textbf{0.09$\pm$0.04}\\
Textual Timing
& 0.12$\pm$0.11 & 0.53$\pm$0.04 & \textbf{0.03$\pm$0.00}
& 0.00$\pm$0.00 & 0.41$\pm$0.02 & 0.12$\pm$0.03\\
FreeAudio
& 0.08$\pm$0.11 & 0.47$\pm$0.06 & 0.33$\pm$0.18
& 0.00$\pm$0.00 & 0.22$\pm$0.19 & 0.96$\pm$0.04\\
Prompt Relay
& 0.28$\pm$0.18 & 0.74$\pm$0.12 & 0.12$\pm$0.08
& 0.24$\pm$0.17 & 0.62$\pm$0.04 & 0.38$\pm$0.06\\
\rowcolor{gray!10}
TimeSteer (Ours)
& \textbf{0.64$\pm$0.09} & \textbf{0.90$\pm$0.01} & \textbf{0.03$\pm$0.00}
& \textbf{0.64$\pm$0.09} & \textbf{0.81$\pm$0.15} & \textbf{0.09$\pm$0.07}\\
\midrule
\multicolumn{7}{l}{\textit{Single speaker, multiple utterances}}\\
Uncontrolled
& 0.30$\pm$0.17 & 0.74$\pm$0.06 & 0.07$\pm$0.11
& 0.20$\pm$0.00 & 0.81$\pm$0.01 & \textbf{0.04$\pm$0.01}\\
Textual Timing
& 0.28$\pm$0.08 & 0.77$\pm$0.05 & \textbf{0.03$\pm$0.03}
& 0.10$\pm$0.07 & 0.72$\pm$0.03 & 0.08$\pm$0.03\\
FreeAudio
& 0.12$\pm$0.08 & 0.53$\pm$0.24 & 0.45$\pm$0.25
& 0.00$\pm$0.00 & 0.11$\pm$0.07 & 0.97$\pm$0.03\\
Prompt Relay
& 0.40$\pm$0.19 & 0.78$\pm$0.04 & 0.09$\pm$0.06
& 0.50$\pm$0.00 & \textbf{0.88$\pm$0.00} & 0.14$\pm$0.01\\
\rowcolor{gray!10}
TimeSteer (Ours)
& \textbf{0.90$\pm$0.12} & \textbf{0.87$\pm$0.08} & 0.06$\pm$0.09
& \textbf{0.82$\pm$0.08} & \textbf{0.88$\pm$0.03} & 0.05$\pm$0.03\\
\midrule
\multicolumn{7}{l}{\textit{Multiple speakers, single utterance}}\\
Uncontrolled
& 0.20$\pm$0.00 & 0.51$\pm$0.05 & \textbf{0.03$\pm$0.01}
& 0.08$\pm$0.11 & 0.58$\pm$0.10 & \textbf{0.10$\pm$0.05}\\
Textual Timing
& 0.16$\pm$0.09 & 0.51$\pm$0.03 & \textbf{0.03$\pm$0.01}
& 0.00$\pm$0.00 & 0.47$\pm$0.02 & \textbf{0.10$\pm$0.04}\\
FreeAudio
& 0.12$\pm$0.18 & 0.41$\pm$0.03 & 0.21$\pm$0.08
& 0.00$\pm$0.00 & 0.15$\pm$0.08 & 0.93$\pm$0.05\\
Prompt Relay
& 0.28$\pm$0.18 & 0.56$\pm$0.10 & 0.36$\pm$0.06
& 0.16$\pm$0.26 & 0.69$\pm$0.06 & 0.32$\pm$0.09\\
\rowcolor{gray!10}
TimeSteer (Ours)
& \textbf{0.52$\pm$0.23} & \textbf{0.82$\pm$0.05} & 0.07$\pm$0.06
& \textbf{0.56$\pm$0.09} & \textbf{0.71$\pm$0.15} & 0.11$\pm$0.05\\
\midrule
\multicolumn{7}{l}{\textit{Multiple speakers, multiple utterances}}\\
Uncontrolled
& 0.20$\pm$0.10 & 0.71$\pm$0.07 & \textbf{0.09$\pm$0.04}
& 0.08$\pm$0.04 & 0.69$\pm$0.02 & \textbf{0.07$\pm$0.02}\\
Textual Timing
& 0.16$\pm$0.09 & 0.67$\pm$0.09 & 0.10$\pm$0.09
& 0.08$\pm$0.04 & 0.64$\pm$0.03 & 0.09$\pm$0.06\\
FreeAudio
& 0.16$\pm$0.11 & 0.52$\pm$0.09 & 0.48$\pm$0.07
& 0.02$\pm$0.04 & 0.08$\pm$0.06 & 0.98$\pm$0.01\\
Prompt Relay
& 0.26$\pm$0.05 & 0.66$\pm$0.10 & 0.24$\pm$0.10
& 0.24$\pm$0.09 & 0.78$\pm$0.02 & 0.20$\pm$0.04\\
\rowcolor{gray!10}
TimeSteer (Ours)
& \textbf{0.66$\pm$0.09} & \textbf{0.86$\pm$0.06} & \textbf{0.09$\pm$0.07}
& \textbf{0.66$\pm$0.09} & \textbf{0.81$\pm$0.07} & 0.10$\pm$0.04\\
\bottomrule
\end{tabular*}
\caption{Cross-seed robustness across speaking structures. Values are mean$\pm$std over five seeds. Best mean results within each backbone and structure are in bold.}
\label{tab:structure-cross-seed}
\end{table*}

\subsubsection{Cross-Metric Consistency and Candidate Selection}
\label{app:cross-metric-consistency}

\begin{figure*}[ht]
\centering
\includegraphics[width=\linewidth]{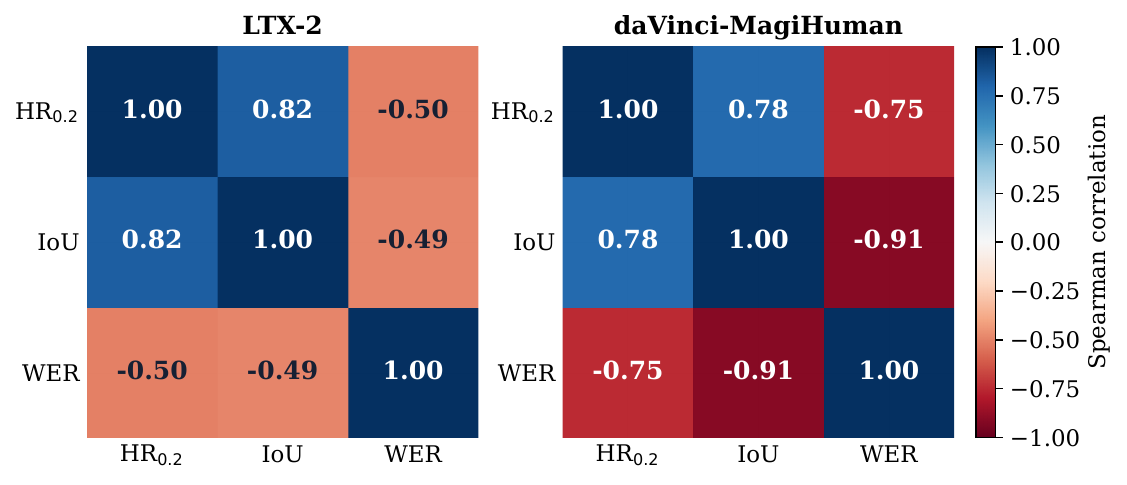}
\caption{Sample-level Spearman correlations among HR$_{0.2}$, IoU, and WER for TimeSteer across five seeds. Positive HR$_{0.2}$--IoU correlations indicate agreement between the temporal metrics, while negative correlations with WER indicate that better temporal control tends to coincide with lower transcription error.}
\label{fig:metric-consistency-correlation}
\end{figure*}

We compute Spearman correlations among HR$_{0.2}$, IoU, and WER across individual generations to examine whether accurate temporal control coincides with faithful speech generation. Figure~\ref{fig:metric-consistency-correlation} shows strong agreement between HR$_{0.2}$ and IoU on both backbones. Their correlations with WER are consistently negative, indicating that improved interval placement generally accompanies, rather than compromises, transcription fidelity.

This relationship is particularly useful for candidate selection. Among outputs with HR$_{0.2}=1$ and IoU$\geq0.8$, approximately $90\%$ also achieve WER$\leq0.1$ on both backbones. Temporal and transcription metrics can therefore be used jointly to identify generations that satisfy both timing and content requirements.

\subsubsection{Additional Ablation Studies}
\label{app:additional-ablations}

We provide additional analyses of both stages of TimeSteer. We first examine source-localization behavior across denoising steps and its sensitivity to the localization threshold, and then study how the denoising-step schedule used for Region-Aware Latent Remapping affects performance.

\begin{figure*}[ht]
\centering
\includegraphics[width=\linewidth]{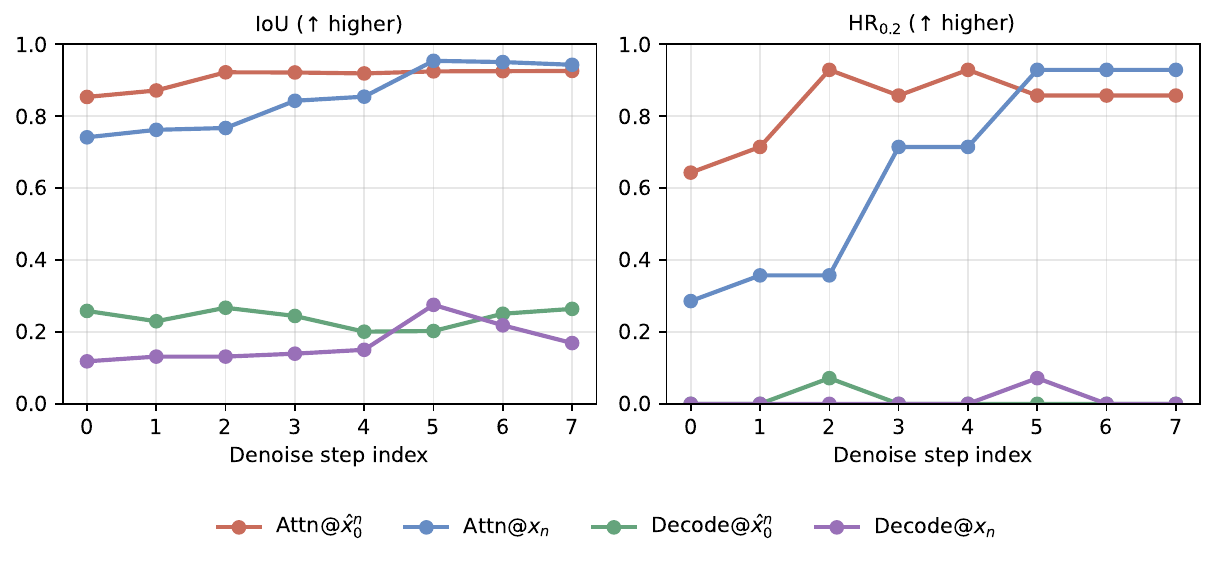}
\caption{\textbf{Step-wise extension of the source-localization ablation.} Left: mean IoU. Right: HR$_{0.2}$. Curves use the same evaluation data as the aggregated comparison in the main paper.}
\label{fig:src-localization-by-step}
\end{figure*}

\paragraph{Source localization across denoising steps.}
The main paper compares four source-localization strategies using results aggregated over denoising. Figure~\ref{fig:src-localization-by-step} disaggregates the same ablation data by denoising step. Attn@$\hat{x}_0^n$ reaches high IoU and HR$_{0.2}$ within the first few steps and remains stable, whereas Attn@$x_n$ becomes competitive only near the end. Both decode-based variants remain substantially less accurate. This step-wise view confirms that the clean estimate exposes reliable timing before it is recoverable from intermediate audio.

\begin{figure*}[ht]
\centering
\includegraphics[width=\linewidth]{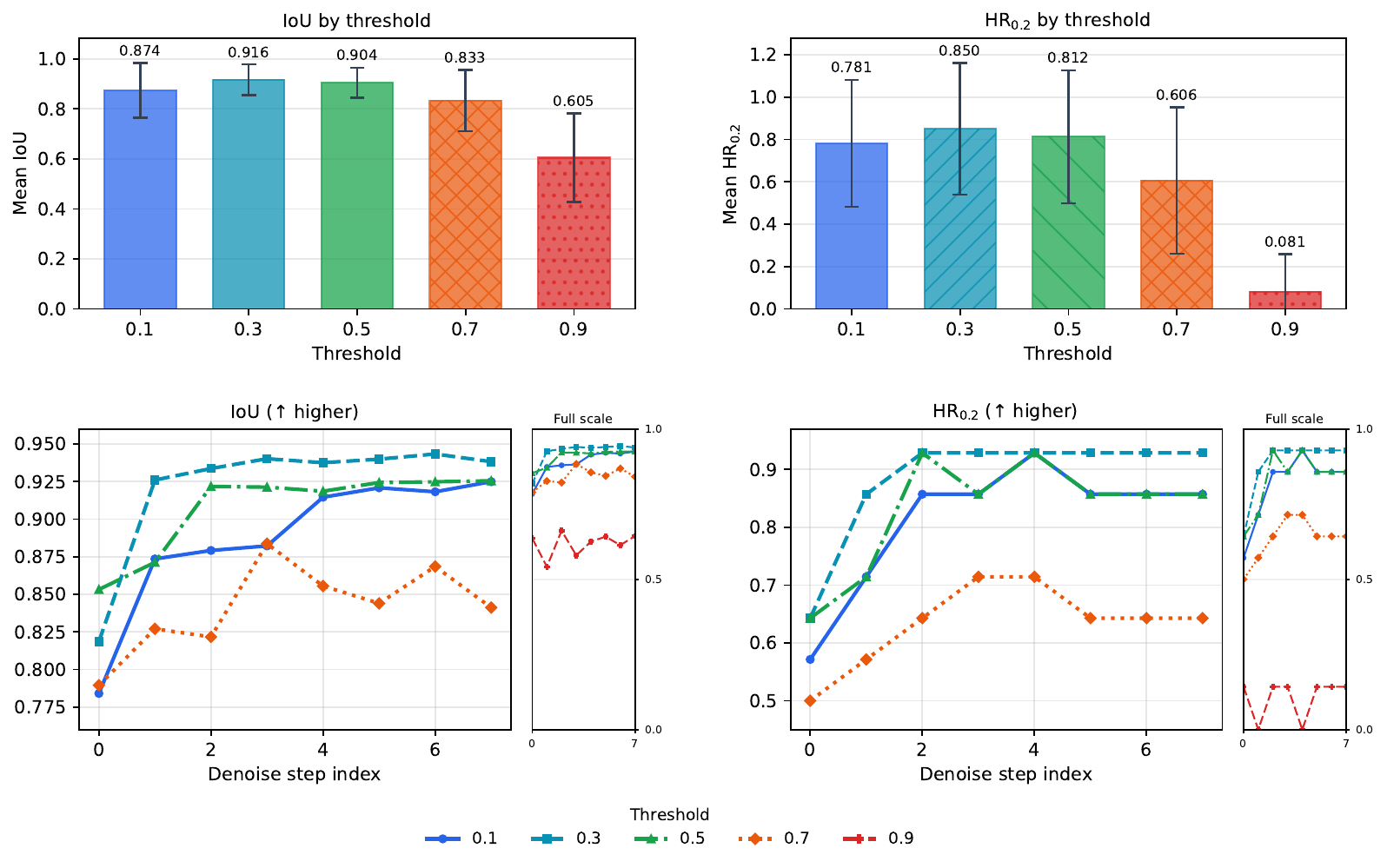}
\caption{\textbf{Sensitivity to the localization threshold for Attn@$\hat{x}_0^n$.} Left: IoU. Right: HR$_{0.2}$. Top: mean over all denoising steps, with values labeled above the bars and error bars showing standard deviation across cases. Bottom: step-wise means; the narrow panels show the full $[0,1]$ scale. Colors and line styles identify the threshold.}
\label{fig:src-localization-threshold-sweep}
\end{figure*}

\paragraph{Localization-threshold sensitivity.}
We vary the relative attention-mass threshold that converts the Attn@$\hat{x}_0^n$ mass into a source span. Figure~\ref{fig:src-localization-threshold-sweep} shows a broad optimum around $0.3$--$0.5$: threshold $0.3$ gives the highest aggregate IoU and HR$_{0.2}$, while the operating value $0.5$ remains close across denoising steps. Larger thresholds contract the detected span; at $0.9$, both metrics drop sharply.

\begin{figure*}[ht]
\centering
\includegraphics[width=\linewidth]{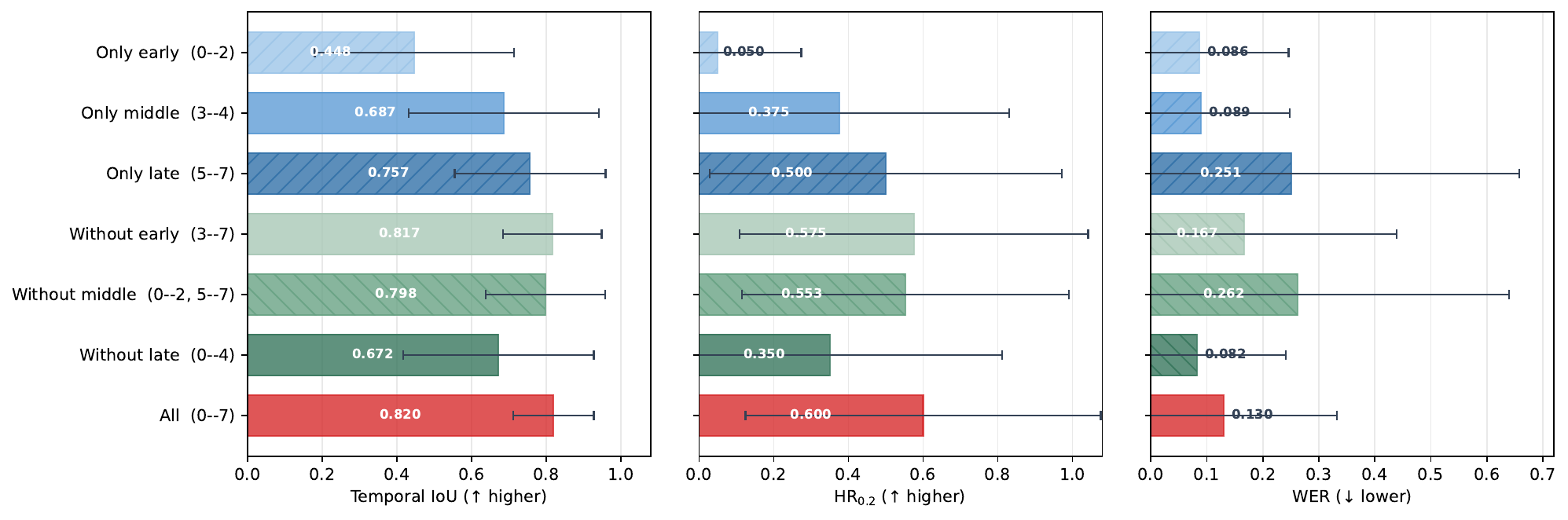}
\caption{\textbf{Sensitivity to the denoising-step schedule for Region-Aware Latent Remapping.} Early, middle, and late denote steps $0$--$2$, $3$--$4$, and $5$--$7$, respectively. From left to right: temporal IoU, HR$_{0.2}$, and WER. Bars show means over the fixed stratified 20-case subset; error bars show standard deviation across cases. The all-step schedule is highlighted in red.}
\label{fig:warp-denoise-step-ablation}
\end{figure*}

\paragraph{Remapping-schedule sensitivity.}
We next vary the denoising steps at which Region-Aware Latent Remapping is applied. We partition the eight steps into early ($0$--$2$), middle ($3$--$4$), and late ($5$--$7$) stages. Because eight steps cannot be divided equally into three contiguous stages, the middle stage contains two steps, whereas the early and late stages contain three each. We evaluate both stage-only schedules and their complementary stage-omission schedules on the same deterministic 20-case subset, comprising the first five benchmark entries from each speaking structure. Figure~\ref{fig:warp-denoise-step-ablation} shows that applying remapping at all steps gives the best IoU and HR$_{0.2}$ while maintaining a low WER. The late stage is particularly important: using only late steps remains substantially stronger than using only early or middle steps, while omitting late steps causes the largest degradation in scheduling accuracy. By contrast, removing either the early or middle stage retains most of the all-step remapping performance.

\subsubsection{Additional Qualitative Comparison}
\label{app:qualitative-results}
\begin{figure*}[ht]
    \centering
    \includegraphics[width=\linewidth]{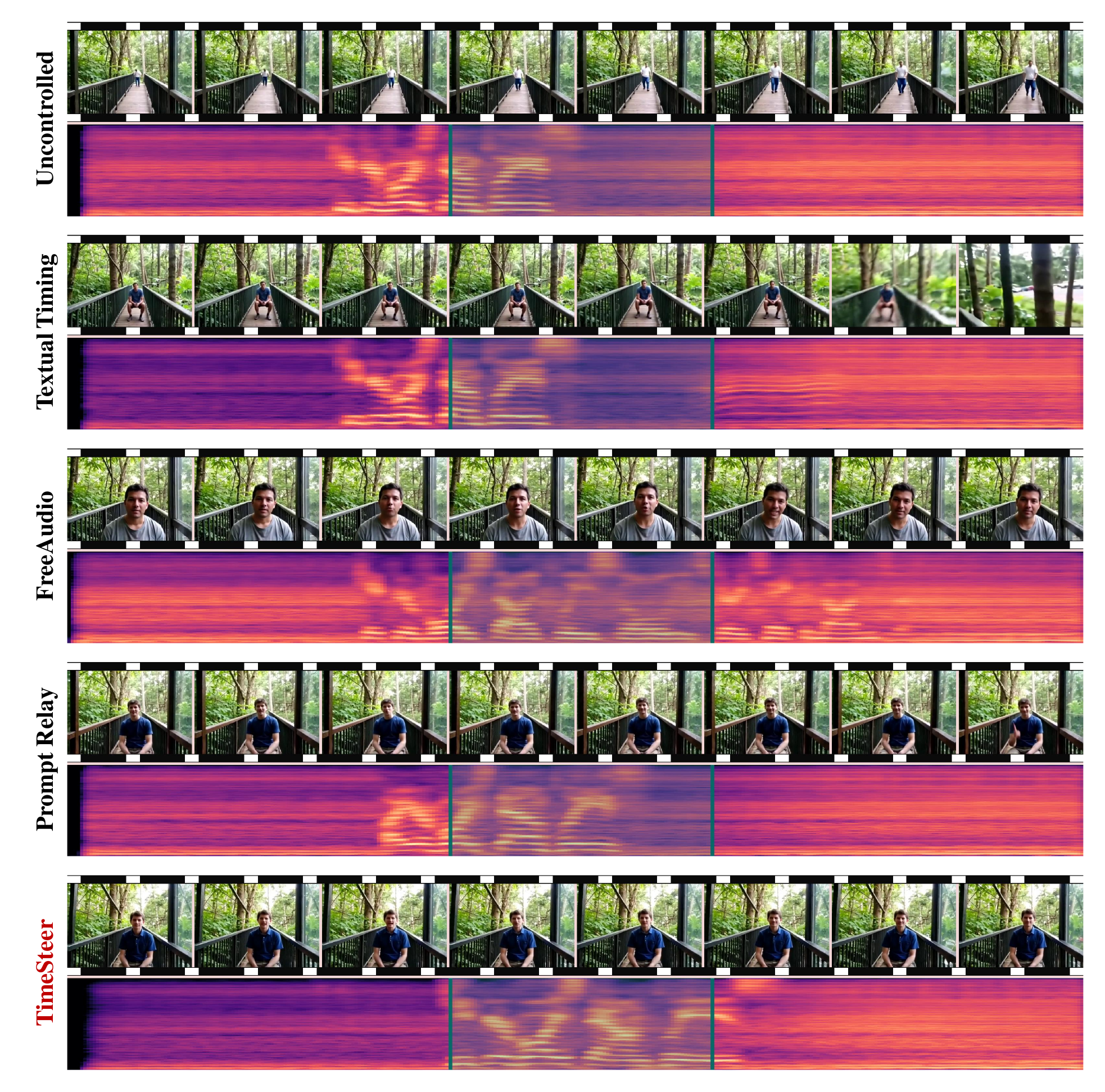}
    \caption{A sample generated by LTX-2. The prompt is "On a rainforest boardwalk trail with dense foliage, a male host sits upright addressing the viewer and says: `keep your voice down please'. Traffic noise and honking continue outside the window." The target interval is [1.9s, 3.2s].}
    \label{fig:ltx2-single-utterance}
\end{figure*}

\begin{figure*}[ht]
    \centering
    \includegraphics[width=\linewidth]{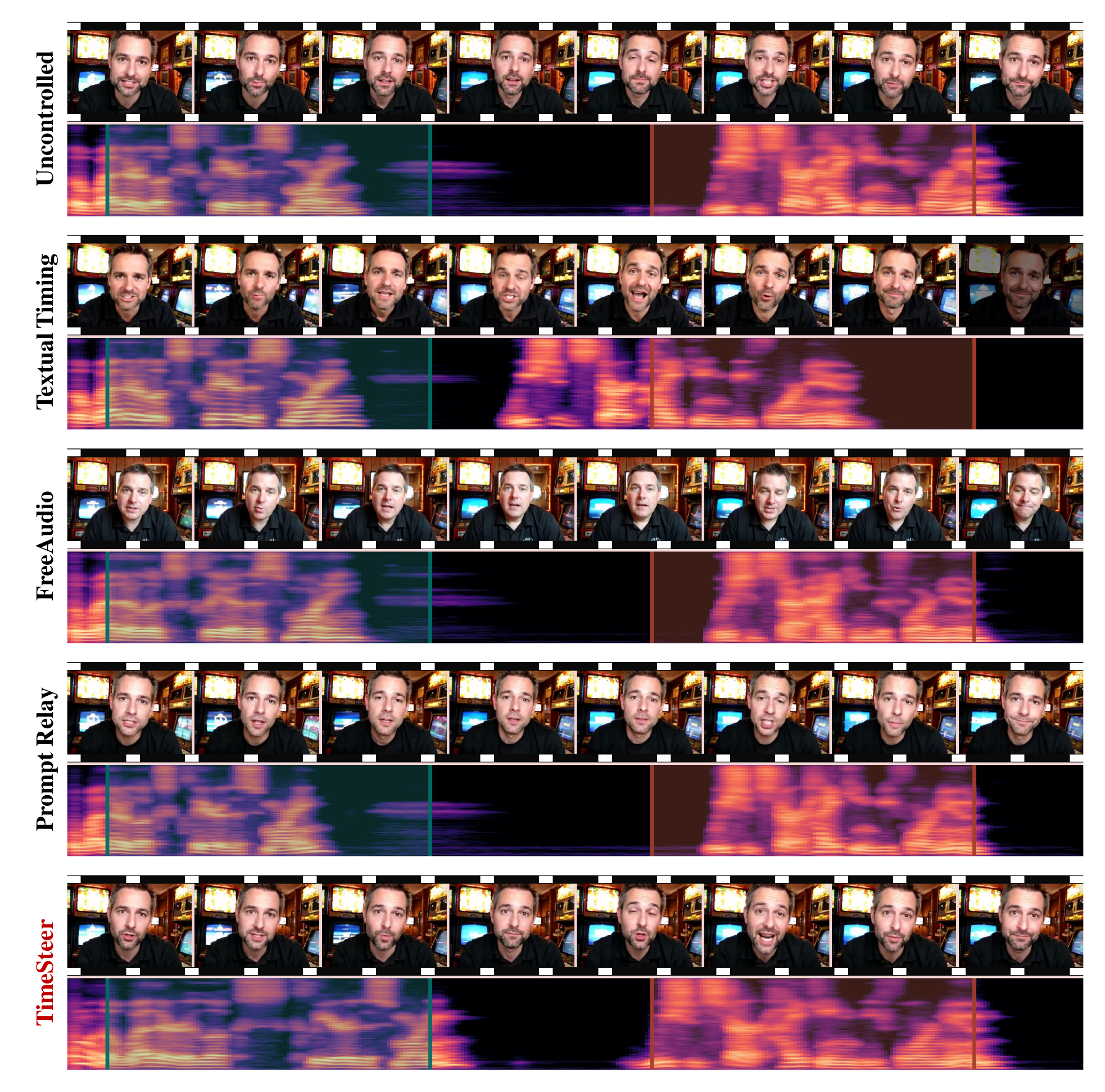}
    \caption{A sample generated by LTX-2. The prompt is "In a retro arcade with flashing game cabinets, a male cashier faces the camera and says: `the weather looks clear this morning'. Following a noticeable pause, he says: `we should start hiking before noon'. The room stays quiet afterward." The target intervals are [0.2s, 1.8s] and [2.9s, 4.5s].}
    \label{fig:ltx2-multiple-utterances}
\end{figure*}

\begin{figure*}[ht]
    \centering
    \includegraphics[width=\linewidth]{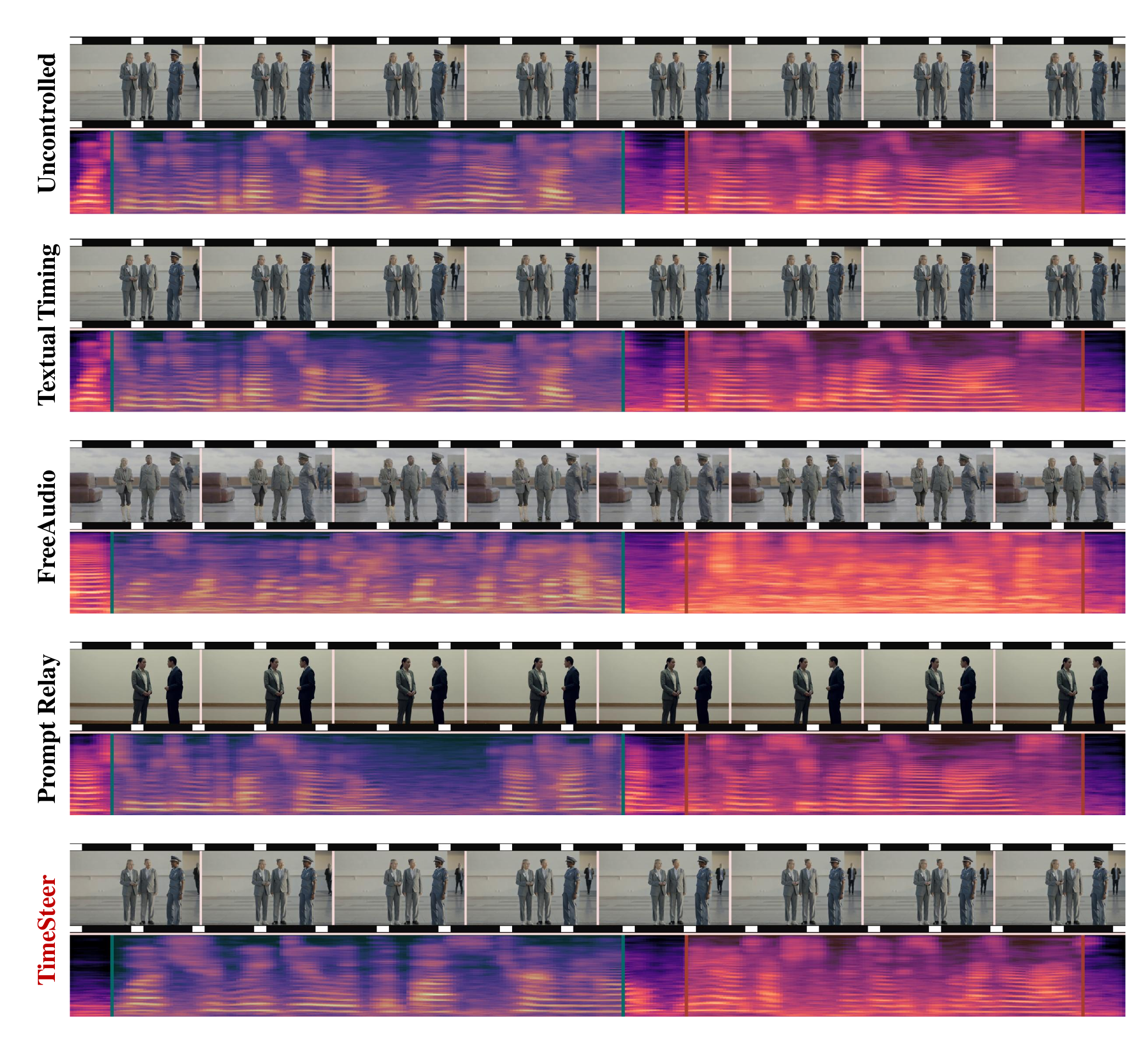}
    \caption{A sample generated by daVinci-MagiHuman. The prompt is "During the afternoon at the bus terminal, a young female station attendant in a charcoal blazer speaks with an older female transportation worker in a neatly pressed uniform. The first person asks: `what should we do with the bus terminal'. The second responds: `leave it untouched until this supervisor arrives'. Elsewhere around the bus terminal, only faint room tone remains after the line." The target intervals are [0.2s, 2.6s] and [2.9s, 4.8s].}
    \label{fig:davinci-multiple-speakers-bus-terminal}
\end{figure*}

\begin{figure*}[ht]
    \centering
    \includegraphics[width=\linewidth]{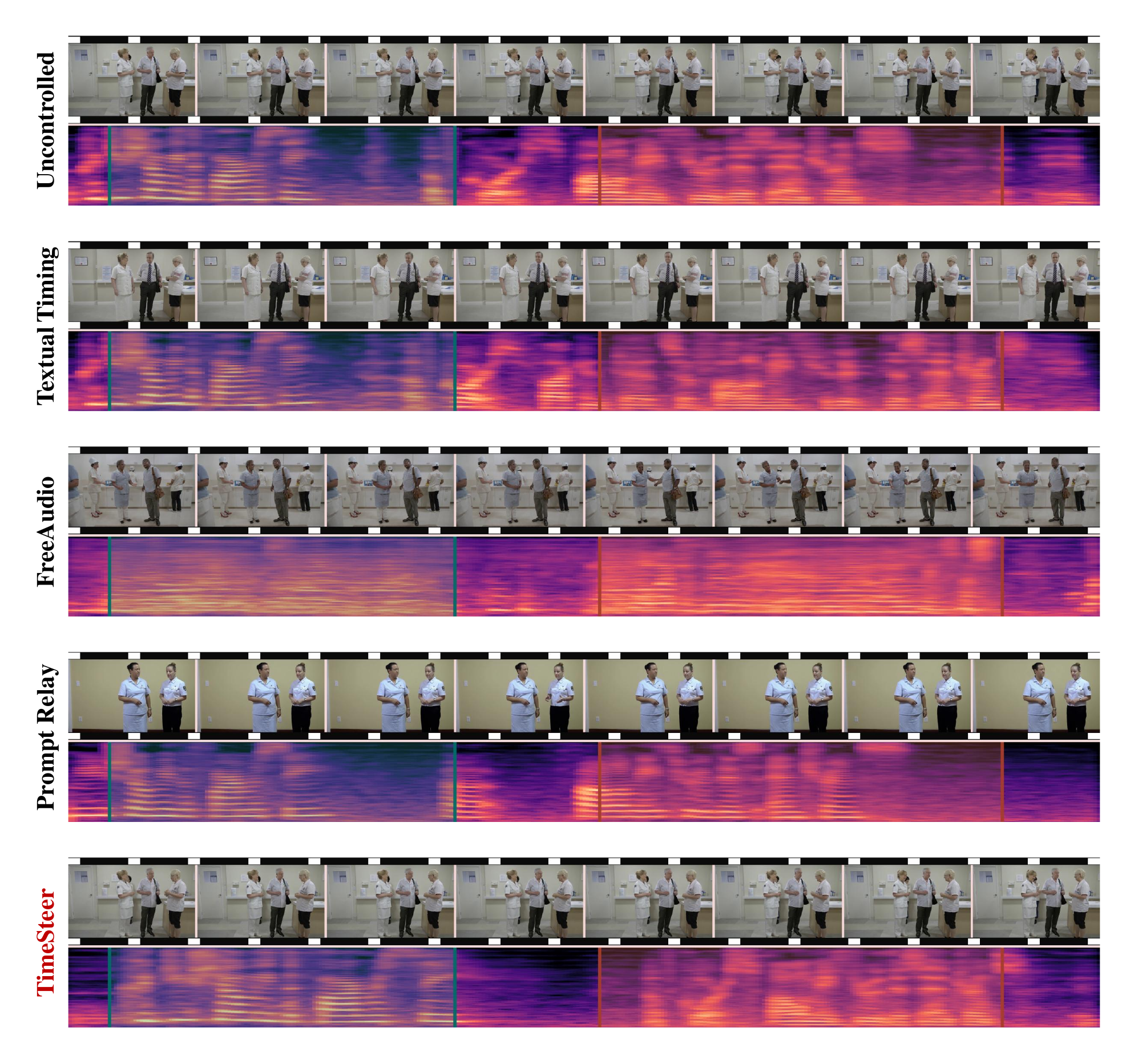}
    \caption{A sample generated by daVinci-MagiHuman. The prompt is "Alongside the pharmacy, a middle-aged female clinic receptionist in a neatly pressed uniform speaks with a middle-aged male dental assistant with a small shoulder bag. The first person asks: 'could you show me the pharmacy'. The second responds: 'of course follow me through this entrance'. Elsewhere around the pharmacy, two staff members discuss a patient quietly behind the speaker." The target intervals are [0.2s, 1.8s] and [2.6s, 4.5s].}
    \label{fig:davinci-multiple-speakers-pharmacy}
\end{figure*}

Figures~\ref{fig:ltx2-single-utterance}--\ref{fig:davinci-multiple-speakers-pharmacy} present qualitative comparisons among the five methods, with two examples from LTX-2 and two from daVinci-MagiHuman. For each method, the upper row shows sampled video frames and the lower row shows the temporally aligned mel-spectrogram; shaded regions indicate the target speech intervals.
Across both backbones, TimeSteer places the speech activity within the specified target intervals more accurately than the competing methods. At the same time, it preserves the acoustic context outside the controlled intervals, including
the background sounds specified by the prompt, rather than suppressing or displacing them together with the speech. The corresponding video frames remain visually coherent and exhibit no prominent artifacts caused by the temporal remapping in these examples.


\end{document}